\documentclass{article}

\usepackage{arxiv}
\usepackage{tcolorbox}

\usepackage[utf8]{inputenc} 
\usepackage[T1]{fontenc}    
\usepackage{hyperref}       
\usepackage{url}            
\usepackage{booktabs}       
\usepackage{amsfonts}       
\usepackage{nicefrac}       
\usepackage{microtype}      
\usepackage{xcolor}         
\usepackage{tikz}
\usetikzlibrary{calc,decorations.pathreplacing}  

\usepackage{amsmath}
\usepackage{amsthm}

\ifthenelse{\isundefined{\definition}}{\newtheorem{definition}{Definition}}{}
\ifthenelse{\isundefined{\assumption}}{\newtheorem{assumption}{Assumption}}{}
\ifthenelse{\isundefined{\hypothesis}}{\newtheorem{hypothesis}{Hypothesis}}{}
\ifthenelse{\isundefined{\proposition}}{\newtheorem{proposition}{Proposition}}{}
\ifthenelse{\isundefined{\theorem}}{\newtheorem{theorem}{Theorem}}{}
\ifthenelse{\isundefined{\lemma}}{\newtheorem{lemma}{Lemma}}{}
\ifthenelse{\isundefined{\corollary}}{\newtheorem{corollary}{Corollary}}{}
\ifthenelse{\isundefined{\alg}}{\newtheorem{alg}{Algorithm}}{}
\ifthenelse{\isundefined{\example}}{\newtheorem{example}{Example}}{}

\newcommand{\contribicon}{\includegraphics[height=1.5ex]{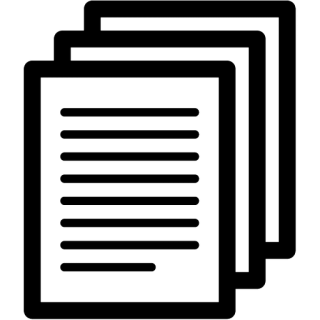}}
\newcommand{\equalcontrib}{\raisebox{0.6ex}{\contribicon}}

\newcommand{\trainalg}{\mathcal{A}}

\definecolor{customred}{HTML}{e74c3c}
\definecolor{custompurple}{HTML}{7030A0}
\definecolor{customblue}{HTML}{45BFC6}
\definecolor{customgold}{HTML}{D4AF37}

\usepackage{subcaption}    
\usepackage{enumitem}

\usepackage{multirow}
\usepackage{adjustbox} 
\usepackage{booktabs}
\usepackage{makecell}

\title{Data-efficient pre-training by scaling synthetic megadocs}

\author{%
  Konwoo Kim\equalcontrib, Suhas Kotha\equalcontrib \\ Yejin Choi, Tatsunori Hashimoto, Nick Haber, Percy Liang \\
  Stanford University \\
}

\begin{document}

\maketitle

\renewcommand{\thefootnote}{}
\footnotetext[1]{{\contribicon\;Equal contribution. Random advisor order.}}
\renewcommand{\thefootnote}{\arabic{footnote}}

\begin{abstract}
\noindent Synthetic data augmentation has emerged as a promising solution when pre-training is constrained by data rather than compute. We study how to design synthetic data algorithms that achieve better loss scaling: not only lowering loss at finite compute but especially as compute approaches infinity.
We first show that pre-training on web data mixed with synthetically generated rephrases improves i.i.d.~validation loss on the web data, despite the synthetic data coming from an entirely different distribution. 
With optimal mixing and epoching, loss and benchmark accuracy improve without overfitting as the number of synthetic generations grows, plateauing near $1.48\times$ data efficiency at 32 rephrases per document.
We find even better loss scaling under a new perspective: synthetic generations from the same document can form a single substantially longer \emph{megadocument} instead of many short documents. 
We show two ways to construct megadocs: stitching synthetic rephrases from the same web document or stretching a document by inserting rationales. Both methods improve i.i.d.~loss, downstream benchmarks, and especially long-context loss relative to simple rephrasing, increasing data efficiency from $1.48\times$ to $1.80\times$ at $32$ generations per document. Importantly, the improvement of megadocs over simple rephrasing widens as more synthetic data is generated. Our results show how to design synthetic data algorithms that benefit more from increasing compute when data-constrained.

\end{abstract}

\begin{figure}[t]
    \centering
    \begin{minipage}[c]{0.3\linewidth}
        \centering
        \includegraphics[width=\linewidth]{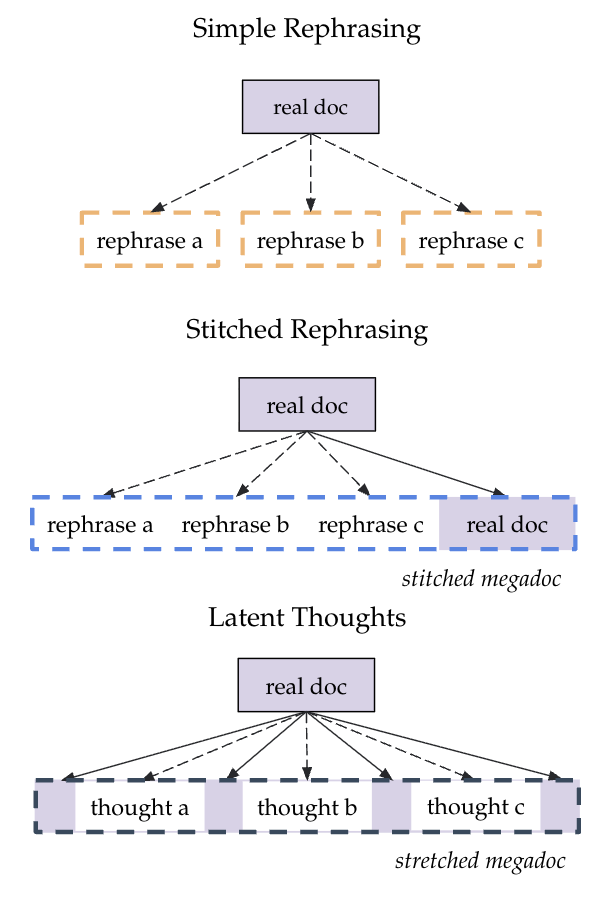}
        
        \phantom{}
    \end{minipage}\hfill
    \begin{minipage}[c]{0.66\linewidth}
        \centering
        \includegraphics[width=\linewidth]{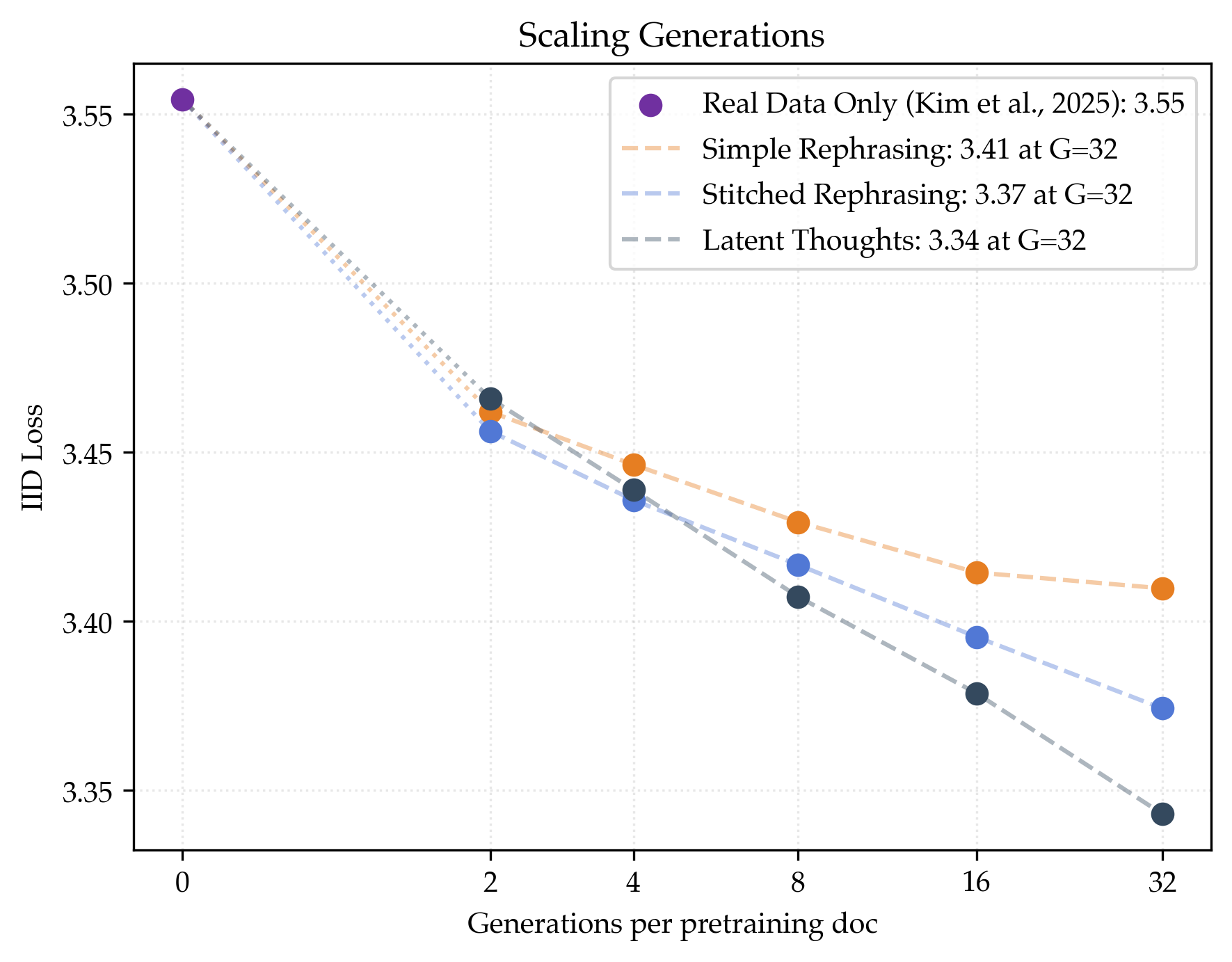}
    \end{minipage}
    \caption{\textbf{Synthetic data by scaling generation count and utilizing megadocs.} When training on 200M unique real tokens, the best 300M parameter model with tuned epoching and regularization achieves $3.55$ loss \citep{kim2025pre}. By mixing synthetically generated rephrases into pre-training, we can improve i.i.d.~loss on the original distribution (orange points), monotonically as the number of generations per document grows. We consider two synthetic data algorithms that leverage multiple generations to produce a single megadoc: stitched rephrasing, which concatenates all generations from the same document (blue points), and latent thoughts, which stretches documents by adding rationales (gray points). Both algorithms improve i.i.d. loss and improve scaling in the number of generations per document. The scaling in synthetic token count is even better as the average latent thought is shorter than the average rephrase. 
    }
    \label{fig:figure-1}
\end{figure}

\section{Introduction}


Compute is growing considerably faster than web text available for language model pre-training \citep{epoch2024trainingcomputeoffrontieraimodelsgrowsby45xperyear,villalobos2024rundatalimitsllm}, motivating the need for data-efficient pre-training. Such recipes aim to better model the underlying distribution of web text, typically by achieving lower loss \citep{kim2025pre,prabhudesai2025diffusionbeatsautoregressivedataconstrained,gladstone2025energybasedtransformersscalablelearners}.
While synthetic data augmentation (e.g. rephrasing \citep{maini2024rephrasingwebrecipecompute}) has emerged as a promising solution, most works focus on its ability to transform the data into different formats that are higher quality \citep{su2025nemotroncctransformingcommoncrawl} or closer to the downstream task \citep{gunasekar2023textbooksneed,liu2023tinygsmachieving80gsm8k}.  
It is underexplored if synthetic data is useful for modeling the original data distribution, the core problem of pre-training.
In this paper, we show how synthetic data can achieve lower i.i.d.~validation loss on web text and we design algorithms that achieve lower loss as the amount of synthetic training data increases. 
We operate under the asymptotically infinite compute regime of \cite{kim2025pre} where we aim to train the best possible 300M parameter model given 200M real tokens of web text under unlimited training compute. Since models at our scale cannot reliably follow instructions, we follow prior work for synthetic data algorithms and assume access to an external synthetic data generator 
(Llama 3.1 8B Instruct) \citep{maini2024rephrasingwebrecipecompute,datologyai2025beyondweblessonsscalingsynthetic,yang2024syntheticcontinuedpretraining}. Although our setting does not pre-train the generator from scratch, ablations and prior work suggest that our synthetic data improvements are not purely artifacts of distillation and could hold when the generator is derived from the same data as the student provided sufficient compute (detailed discussion in Section \ref{sec:stronger-generator}).

We first show how pre-training on real data mixed with rephrases improves i.i.d.~loss on the \emph{original} web distribution 
despite synthetic data coming from a different distribution.
Loss monotonically improves in the number of rephrases per document when carefully tuning mixing and epoching, plateauing without overfitting near $3.41$ i.i.d.~loss at $32$ rephrases relative to the real data only baseline of $3.55$ (Figure \ref{fig:figure-1}). Best estimates of data scaling from \cite{kim2025pre} predict that simple rephrasing needs $1.48\times$ less real data to achieve the same loss as the baseline, marking a significant data efficiency improvement. The loss improvements are matched by downstream benchmarks, improving average accuracy by $5\%$ on PIQA, SciQ, and ARC Easy.


To further improve loss scaling, we introduce the perspective that synthetic generations can form a single long \emph{megadocument} instead of being shuffled as independent short documents. 
We follow this view to propose two methods that generate megadocuments. First, inspired by In-context Pre-training \citep{shi2024incontextpretraininglanguagemodeling}, we propose "stitching" synthetic rephrases generated from the same real document. Second, inspired by Latent Thoughts \citep{ruan2025reasoninglearnlatentthoughts}, we split each real doc into multiple chunks and stretch the document by inserting rationales connecting each prefix and suffix. We illustrate both megadocuments in Figure \ref{fig:figure-1}, left.

Both megadoc algorithms improve over simple rephrasing for i.i.d.~validation loss, long-context loss, and downstream benchmarks. Stitched rephrasing and latent thoughts achieve $1.64\times$ and $1.80\times$ data efficiency at 32 generations, improving upon simple rephrasing's $1.48\times$ data efficiency (Figure \ref{fig:figure-1}). Importantly, megadocs show less signs of plateauing under generation count relative to simple rephrasing. Our loss gains are even more pronounced when measured on validation sets explicitly measuring long-context capabilities such as recent computer science arXiv papers \citep{uncheatable_eval}. The loss improvements are again matched by downstream benchmarks, improving average accuracy by $6\%$ and $9\%$ over the real data only baseline. 

Finally, we show that the benefits of simple rephrasing, stitched rephrasing, and latent thoughts compose with the already data-efficient strategy of ensembling \citep{kim2025pre}. 
We find that an alternative form of synthetic data via self-distillation does not compose with ensembling, demonstrating a qualitatively unique benefit of the synthetic data algorithms scaled in this paper.

We open-source all of our \href{https://wandb.ai/stanford-mercury/suhas-data-efficiency/reports/Data-efficient-pre-training-by-scaling-synthetic-megadocs--VmlldzoxNjIyNjExNA}{runs} on WandB and our \href{https://github.com/marin-community/marin/tree/suhas/data-efficiency}{code} on Github.


\section{Setup}

We are interested in synthetic data algorithms that achieve lower loss given a fixed web dataset and generator model as compute approaches infinity. Since our primary goal is to model the underlying web distribution, we quantify performance by i.i.d.~validation loss on web text, even though the synthetic data comes from a different distribution induced by the generator. To estimate the extent to which we can scale a synthetic data algorithm, we measure how loss changes as a function of the number of synthetic generations and determine where the loss plateaus. We will specifically measure scaling by generating a fixed budget of $G$ generations per document, following training recipes from \cite{kim2025pre} for unlimited compute pre-training with appropriate mixing, and measuring how loss changes as a function of $G$.

\paragraph{Training.} 
As done in \cite{kim2025pre}, we assume access to 200M tokens of web text sourced from 164,000 DCLM documents \citep{li2025datacomplmsearchgenerationtraining} and train over-parameterized 300M parameter autoregressive transformers with 4096 context length and cross-document attention. All of our training runs tune the number of epochs on the real data with appropriate regularization via weight decay. We offer more details on our training algorithm in Appendix \ref{app:training-details}.

\paragraph{Generator.} Throughout this paper, we assume access to an external synthetic data generator of Llama 3.1 8B Instruct. We practically opt for this setting instead of pre-training the generator from scratch since it would be too expensive for us to experiment at data budgets where the generator would be sufficiently capable. Nonetheless, we are optimistic our findings will hold for self-improvement provided sufficient compute and data to train a generator from scratch. This is due to (1) ablations showing how synthetic data helps even more for more capable students, (2) prior work on self-improvement via rephrasing \citep{kotha2025selfimprovement}, inter-document modeling \citep{yang2025syntheticbootstrappedpretraining}, and latent thoughts \citep{zelikman2024quietstarlanguagemodelsteach,hatamizadeh2026rlpreinforcementpretrainingobjective},
and (3) data augmentation in vision helping at almost every scale \citep{steiner2022trainvitdataaugmentation}, even for validation loss \citep{verma2019manifoldmixupbetterrepresentations}. 
We offer detailed discussion and ablations on this design decision in Section \ref{sec:stronger-generator}.



\paragraph{Evaluation.} Following standard pre-training convention, we use validation loss as a proxy for general model capabilities since it scales smoothly and is predictive of downstream benchmarks in \cite{thrush2025improvingpretrainingdatausing,gadre2024languagemodelsscalereliably,chen2025scalinglawspredictingdownstream,kim2025pre} and our experiments. Importantly, we primarily measure loss on a validation set from the original web distribution without mixing synthetic data: we will refer to this as \emph{i.i.d.~loss}. To contextualize the significance of a loss improvement, we estimate data efficiency, or how much additional data the real data baseline needs to match the loss improvement. We estimate data efficiency using the standard recipe data scaling law from \cite{kim2025pre} and defer a detailed explanation to Appendix \ref{app:data-eff-measurement}. In Section \ref{sec:scaling-document-length}, we additionally measure long-context capabilities via loss on naturally long-context data (e.g. recent CS arXiv papers \citep{uncheatable_eval}). Finally, we confirm that our loss improvements translate to better models by measuring average accuracy on the downstream benchmarks of PIQA, SciQ, and ARC Easy, recommended by \cite{thrush2025improvingpretrainingdatausing} as high signal benchmarks at our pre-training scale.
\section{Synthetic data as scaling document count}\label{sec:rephrasing-science}

\subsection{Rephrasing}\label{sec:rephrasing-subsection}


We start with the simplest synthetic data augmentation: rephrasing \citep{maini2024rephrasingwebrecipecompute}. We query the generator to rephrase each of $D$ real docs into an English Wikipedia article (Appendix \ref{app:prompts}) with temperature 1 and a max generation length of 1024 tokens. The average rephrased document is 708 tokens long, shorter than the average DCLM document length of 1243 tokens. To estimate the best possible loss under infinite generation and training compute, we fix a generation compute budget of $G$ generations per document, train the best possible 300M model given unlimited training compute (following the mixing procedure below to epoch real and synthetic data), and measure how loss varies in the number of rephrases. We discuss this scaling design decision in Appendix \ref{app:parameterization}.

\paragraph{Mixing.} After sampling $G$ rephrases per document, we mix two data streams: a real stream with the $D$ real docs and a synthetic stream with the $G \times D$ rephrases and $D$ real docs. \footnote{
Though it may seem redundant to add real data to the synthetic stream in addition to mixing the real stream, it is slightly different since each context window comes from one stream and we are training with cross-document attention. Therefore, including real documents in the synthetic stream results in real and synthetic documents attending to each other. We choose to include the real doc in the synthetic stream since it will be a fair baseline in anticipation of algorithms in Section \ref{sec:scaling-document-length} and it slightly helps loss (Appendix \ref{app:real-in-synthetic}).
}
Each stream is constructed by concatenating a random permutation of documents with EOS tokens. Each training sequence is formed by selecting a stream and consuming enough tokens to fill the next context window, commonly known as ``concat-and-chunk''. Each training batch follows a specified mixing fraction (e.g. mixing fraction $0$ is only real data). In addition to the mixing fraction, we specify the number of epochs to take on the real stream while epoching the synthetic stream indefinitely, determining the total number of training steps.


\subsection{Loss improvement}\label{sec:rephrasing-scaling}

\begin{figure}
    \centering
    \includegraphics[width=0.45\textwidth]{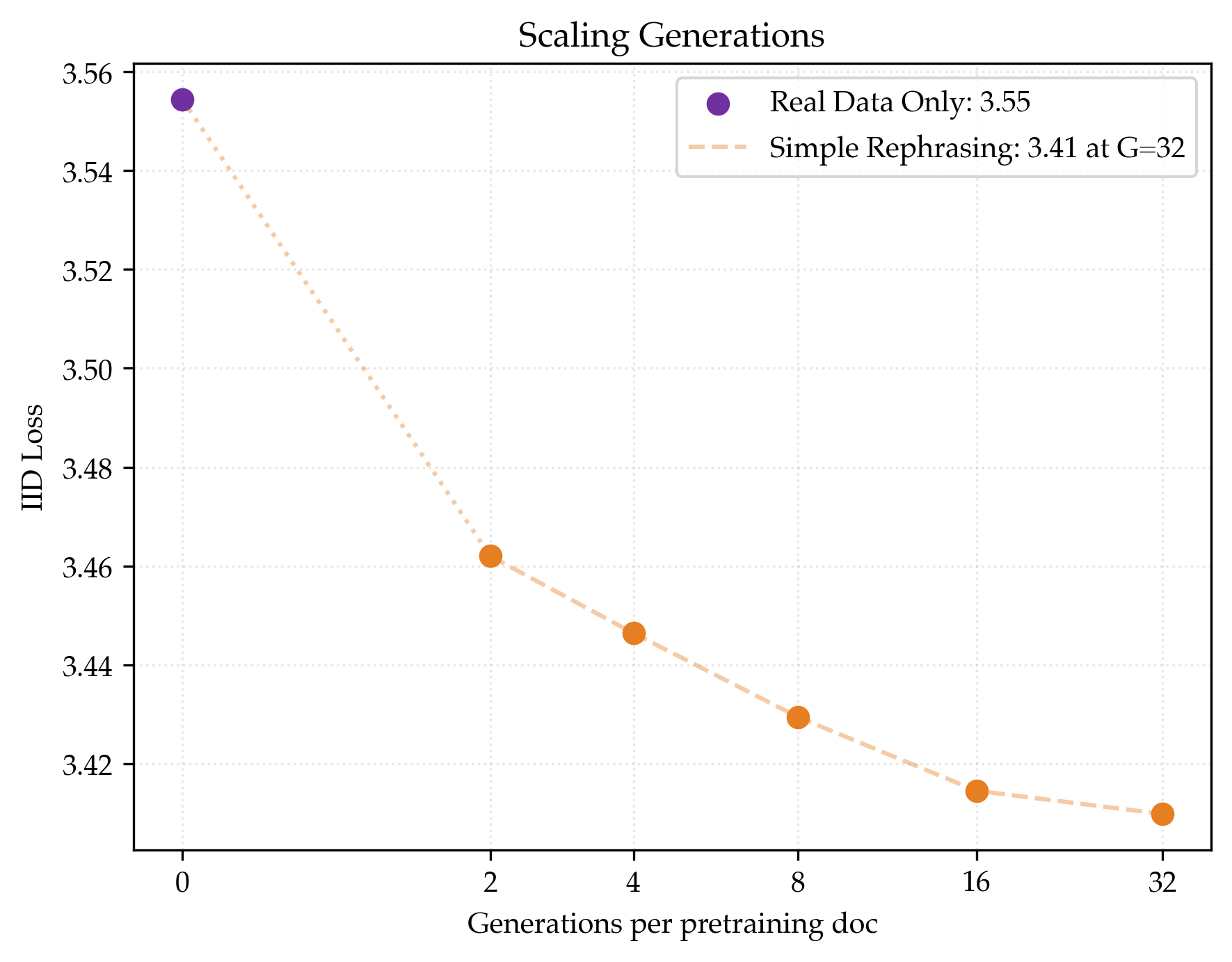}
    \includegraphics[width=0.45\textwidth]{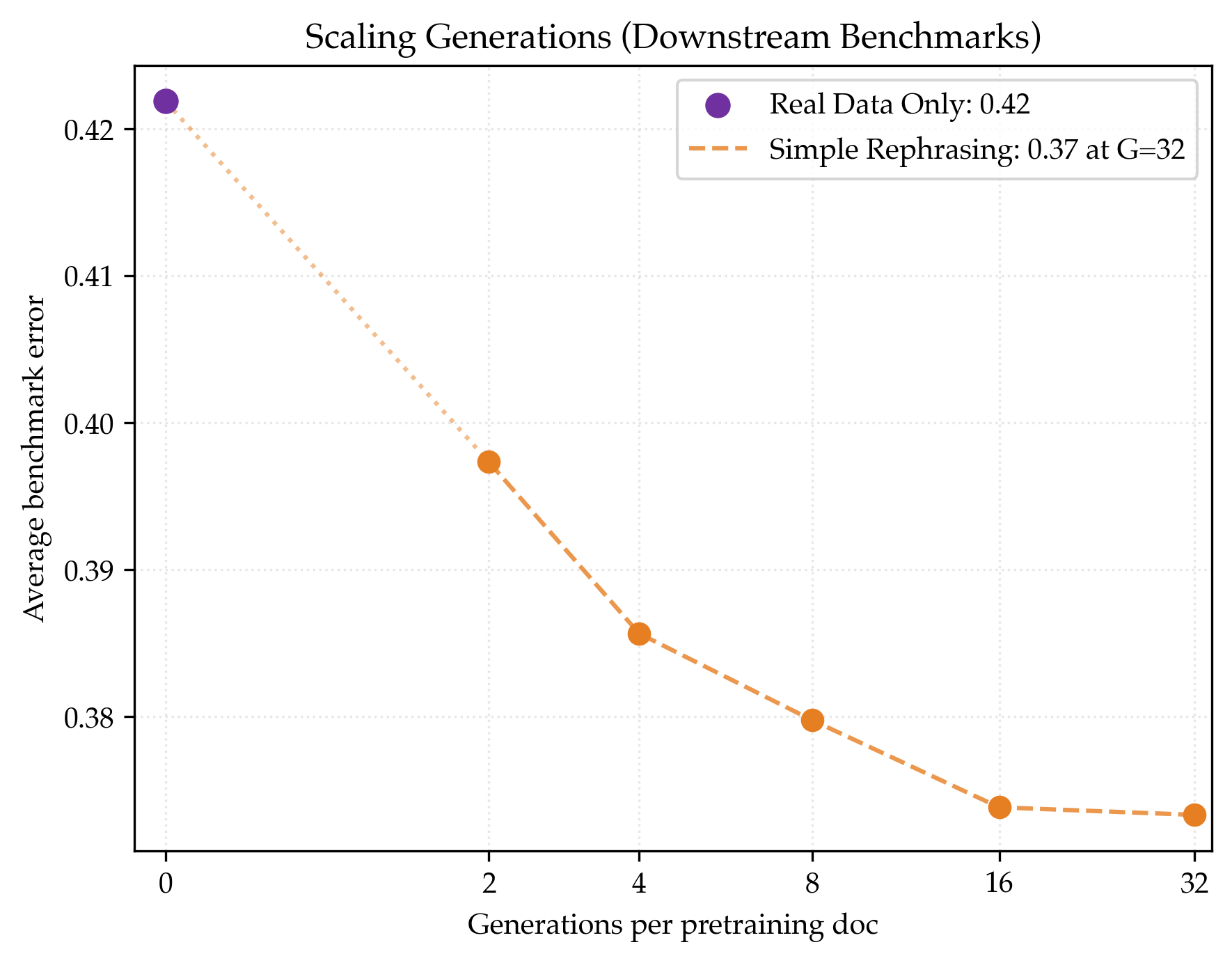}
    \caption{\textbf{Scaling synthetic generations.} Left: Given 200M DCLM tokens, we tune a 300M parameter baseline utilizing epoching and regularization which achieves $3.55$ loss (purple point). We then measure the benefit of sampling $G$ rephrases per real doc and searching for locally optimal learning rate, weight decay, epoch count, and mixing fraction (orange points). We find that loss monotonically improves in the number of rephrases generated, appearing to plateau around $32$ generations at loss $3.41$. Right: The loss improvements and plateau are reflected on downstream benchmarks.}
    \label{fig:shuffled_gen_scaling}
\end{figure}

We find that mixing the synthetic stream offers a significant i.i.d.~loss improvement over our standard pre-training baseline, despite the synthetic data coming from a different distribution than the real data. For our baseline, we pre-train a 300M model on 200M real tokens to get $3.55$ i.i.d.~loss under locally optimal learning rate, epoch count, and weight decay (i.e. incrementing or decrementing any of the hyperparameters does not result in a better model). For rephrasing at a fixed generation count $G$, we additionally search for a locally optimal mixing fraction. Figure \ref{fig:shuffled_gen_scaling}, left shows how simple rephrasing improves i.i.d.~loss over the baseline and scales as the number of generations grows. As we increase the number of generations, the loss monotonically decreases,\footnote{Our mixing parameterization plays a key role in monotonic loss improvement. Specifically, the loss at a given rephrase count can always use the same hyperparameters as a smaller rephrase count while epoching the synthetic stream less. Therefore, a larger rephrase count should never hurt (barring the nuance that we keep a real copy in our synthetic stream). If we did not tune the mixing fraction (e.g. kept a single stream of all synthetic and real docs) we do not expect monotonic scaling since this confuses introducing synthetic data and epoching real data.}, improving loss by $0.14$ and data efficiency by $1.48\times$ at our highest rephrase count of $32$. The loss appears to plateau around $32$ rephrases and will likely not significantly improve with more generations. Furthermore, the loss trends match downstream benchmarks with an accuracy improvement plateauing at $5\%$. We defer a detailed analysis of synthetic data hyperparameter search and interactions to Appendix \ref{app:synth-data-hypers}.


\begin{figure}
    \centering
    \includegraphics[width=0.9\linewidth]{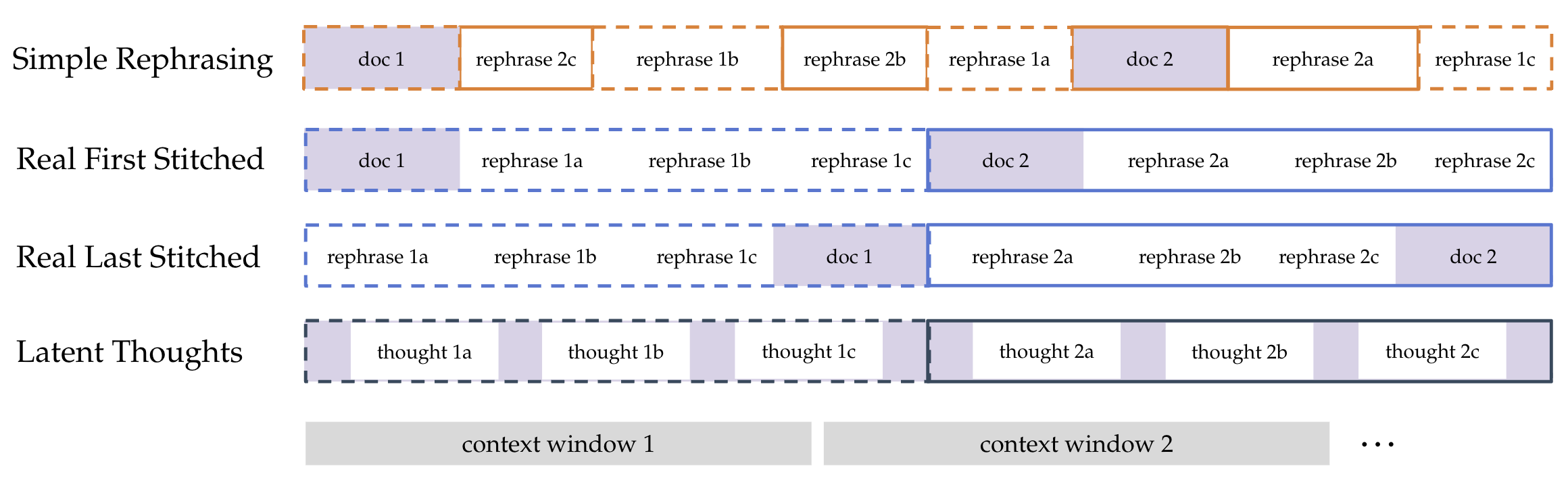}
    \caption{\textbf{Synthetic data streams for megadocs.} We visualize synthetic streams with 2 real docs and 3 generations. Simple rephrasing (Section \ref{sec:rephrasing-science}) permutes all generations and real docs. Stitched rephrasing concatenates all rephrases from the same real doc and either prepends or appends each real doc. Latent thoughts fixes $G$ split points and synthesizes a rationale to derive each suffix from its prefix. Notably, megadocs can exceed the length of the model's context. When training, masking is always disabled across document and megadocument boundaries.  
    }
    \label{fig:sorting_diagram}
\end{figure}

\section{Multiple generations form a megadoc}\label{sec:scaling-document-length}


In the previous section, we showed how synthetic data could lower loss by scaling the total number of documents. However, standard rephrasing subsequently shuffles all of the generations as if they were i.i.d., missing out on the structure that the same real document underlies many related generations. To utilize this information, we consider two synthetic data algorithms that produce one long megadocument instead of multiple short documents.
Our first method is agnostic to generation and applies to any pool of synthetic data (Section \ref{sec:sorting}), while our second method modifies the generation process to expand the real document (Section \ref{sec:latent-thoughts}).
Both methods offer a significant improvement over simple rephrasing for i.i.d.~loss, long-context loss, and downstream benchmarks that scales as the number of generations grows (Section \ref{sec:scaling-gen-count-for-long}).

\subsection{Stitched rephrasing}\label{sec:sorting}

In this section, inspired by In-context Pre-training (ICPT) \citep{shi2024incontextpretraininglanguagemodeling}, we propose the algorithm of ``stitching'' the synthetic stream by concatenating all $G$ rephrases and the real doc (with EOS) for each real doc. Unlike ICPT which requires embedding the entire pre-training corpus and forming a high similarity traversal, stitching synthetic data is essentially free as we know a priori which synthetic generations are related to each other.

When concatenating, we additionally decide whether the real doc goes before or after the $G$ rephrases as visualized in Figure \ref{fig:sorting_diagram}. We note that since the average rephrased document is 708 tokens and the average DCLM document is 1243 tokens, the length of the megadocs will quickly exceed the context window of 4096 tokens.

\begin{figure}
    \centering    
    \includegraphics[width=0.45\textwidth]{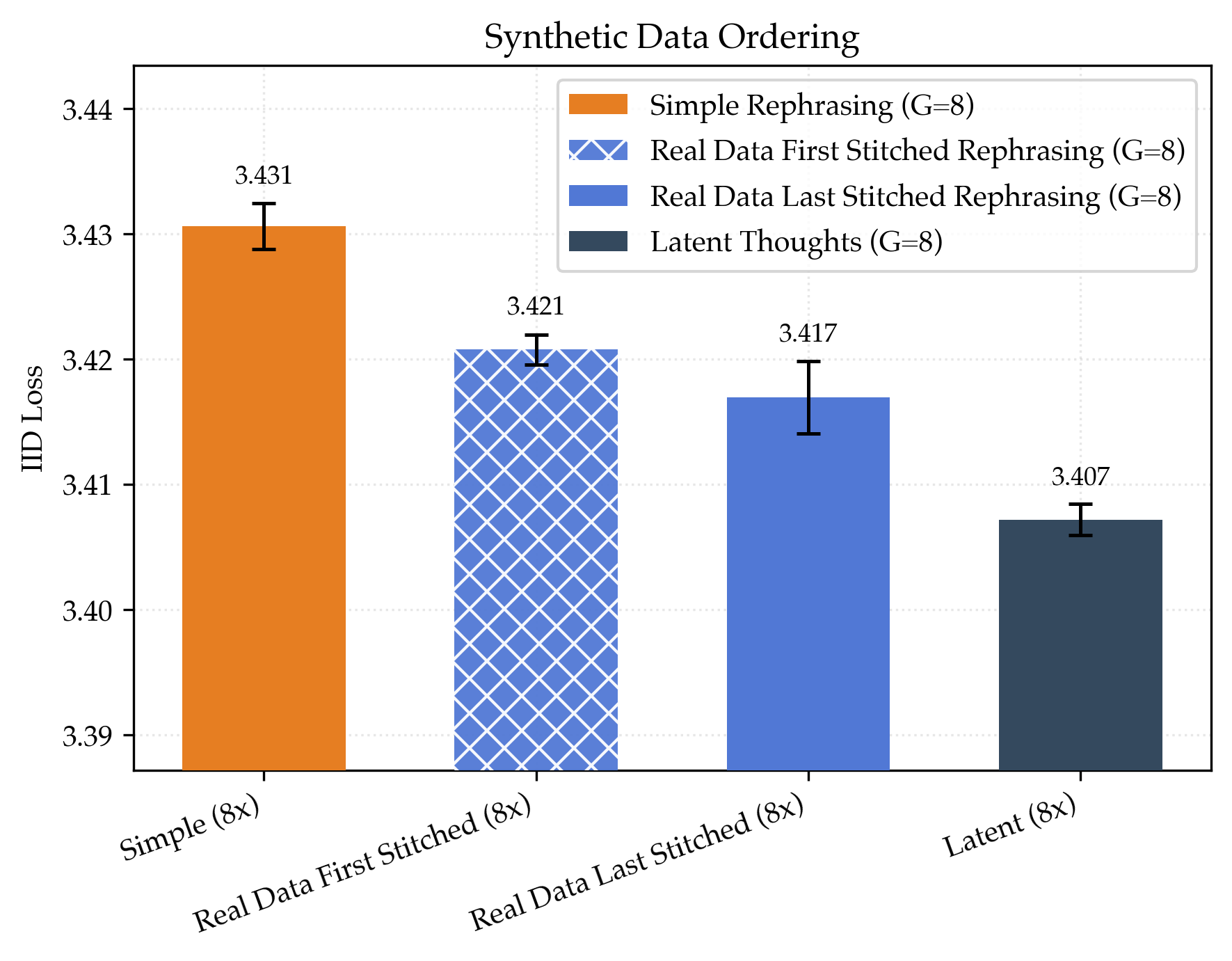}
    \includegraphics[width=0.45\textwidth]{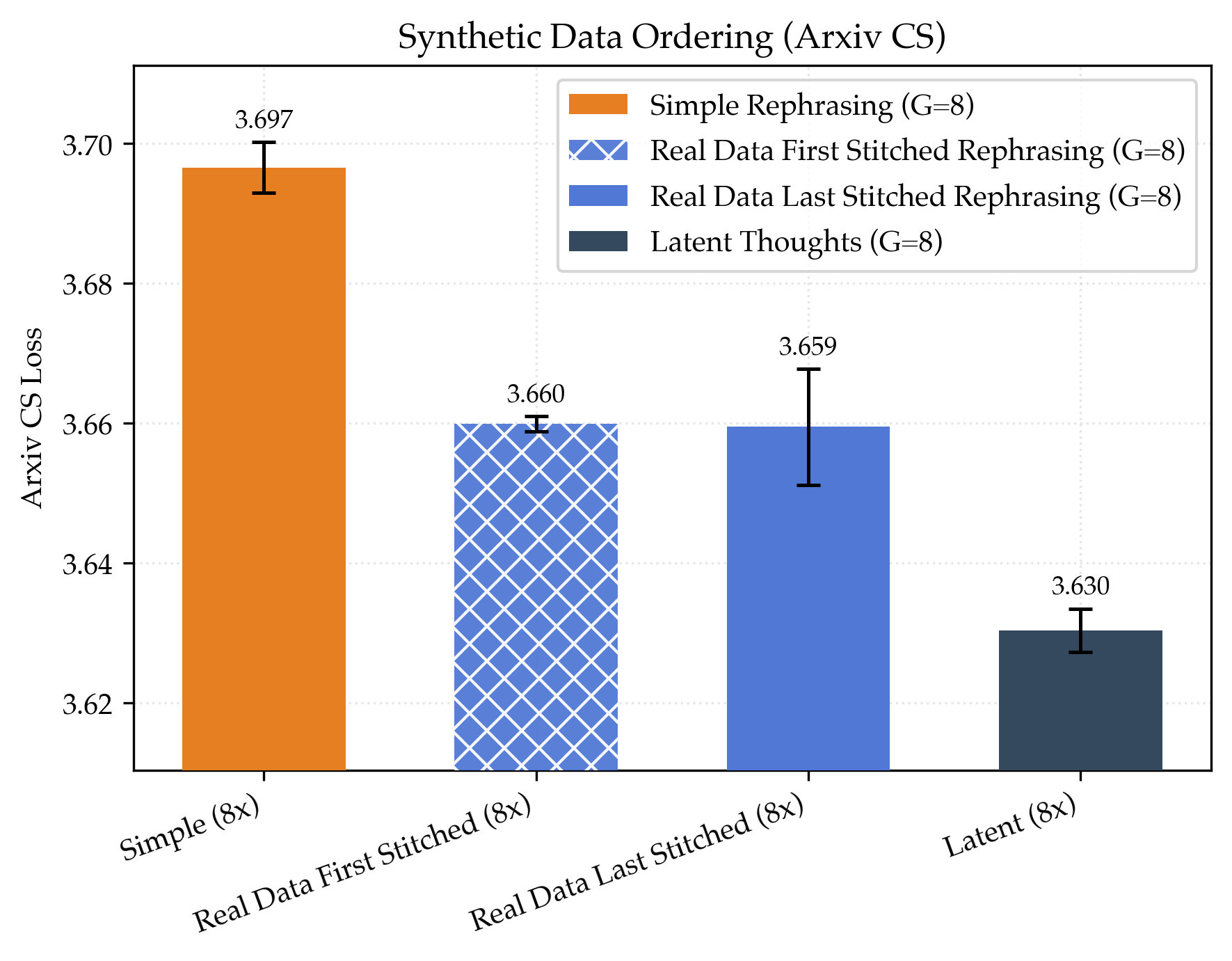}
    \caption{\textbf{Different ways to scale document length.} We compare synthetic data algorithms (visualized in Figure \ref{fig:sorting_diagram}) given 8 generations per pre-training doc. We run 3 seeds of each algorithm. Both variants of stitched rephrasing outperform simple rephrasing with real last slightly outperforming real first. Latent thoughts further improves both i.i.d and arxiv loss.}
    \label{fig:sorting_ablations}
\end{figure}

\paragraph{Stitching improves i.i.d.~loss}
 Figure \ref{fig:sorting_ablations} left shows that stitching with either choice improves i.i.d.~loss over shuffling at 8 generations. This i.i.d.~loss improvement is a previously unknown benefit of stitching, uniquely enabled by mixing in the real data stream (Appendix \ref{app:mixing-iid}).

 \paragraph{Why is stitching the real doc last best?} When stitching, we find that keeping the real doc last outperforms keeping it first (Figure \ref{fig:sorting_ablations} left) and not concatenating it at all (Appendix \ref{app:real-in-synthetic}). This may be related to epiplexity \citep{finzi2026entropyepiplexityrethinkinginformation}, a metric for quantifying the learnable structure in a distribution. One observation of the work is that given a function which is computationally harder to invert than evaluate, learning the inverse can impart more generalizable structure to a learner (best demonstrated by the chess example in their Figures 4, 7). Since the rephrase is generated by a language model and is generally less detailed than the real doc, we suspect that the inverse task of generating the real doc from the rephrase is more valuable for pre-training. 

\paragraph{Stitching improves long-context loss}
We confirm the conventional benefit of stitching on improving long-context modeling by measuring loss on long-context computer science arXiv papers. As shown in Figure \ref{fig:sorting_ablations} right, we observe that stitching offers a larger magnitude loss improvement on the long-context loss relative to the i.i.d.~loss. In Appendix \ref{app:additional-evals}, we show how these improvements hold for additional long-context validation sets.

\subsection{Latent Thoughts}\label{sec:latent-thoughts}


Instead of stitching independent generations to create a megadoc, we can directly generate data that stretches the original document. Inspired by Latent Thoughts \citep{ruan2025reasoninglearnlatentthoughts}, we split the document into $G+1$ equal length pieces and prompt the generator to generate a rationale deriving the suffix from the prefix for all $G$ split points (prompt in Appendix \ref{app:prompts}). We then stretch the original document into a megadoc by inserting the generations wrapped with \texttt{<think></think>} (Figure \ref{fig:sorting_diagram}). Given that the average thought is 424 tokens long, the length of each megadoc will quickly exceed the context window of 4096 tokens. Relative to rephrasing, we are more concerned about distillation effects from a more powerful generator for this augmentation (detailed discussion in \ref{sec:stronger-generator}).

\paragraph{Latent thoughts improves i.i.d.~and long-context loss}
Although \cite{ruan2025reasoninglearnlatentthoughts} failed to observe a loss improvement from latent thoughts on math web data, we show in Figure \ref{fig:sorting_ablations} that the latent thoughts augmentation outperforms simple rephrasing and stitched rephrasing when measured on i.i.d.~loss and long-context loss. In Appendix \ref{app:additional-evals}, we show how these improvements hold for additional validation sets.


\begin{figure}
    \centering

    \begin{minipage}[t]{0.48\textwidth}
        \centering
        \includegraphics[width=\linewidth]{plots/generation_scaling_shuffled_sorted_latents.png}
    \end{minipage}
    \hfill
    \begin{minipage}[t]{0.48\textwidth}
        \centering
        \includegraphics[width=\linewidth]{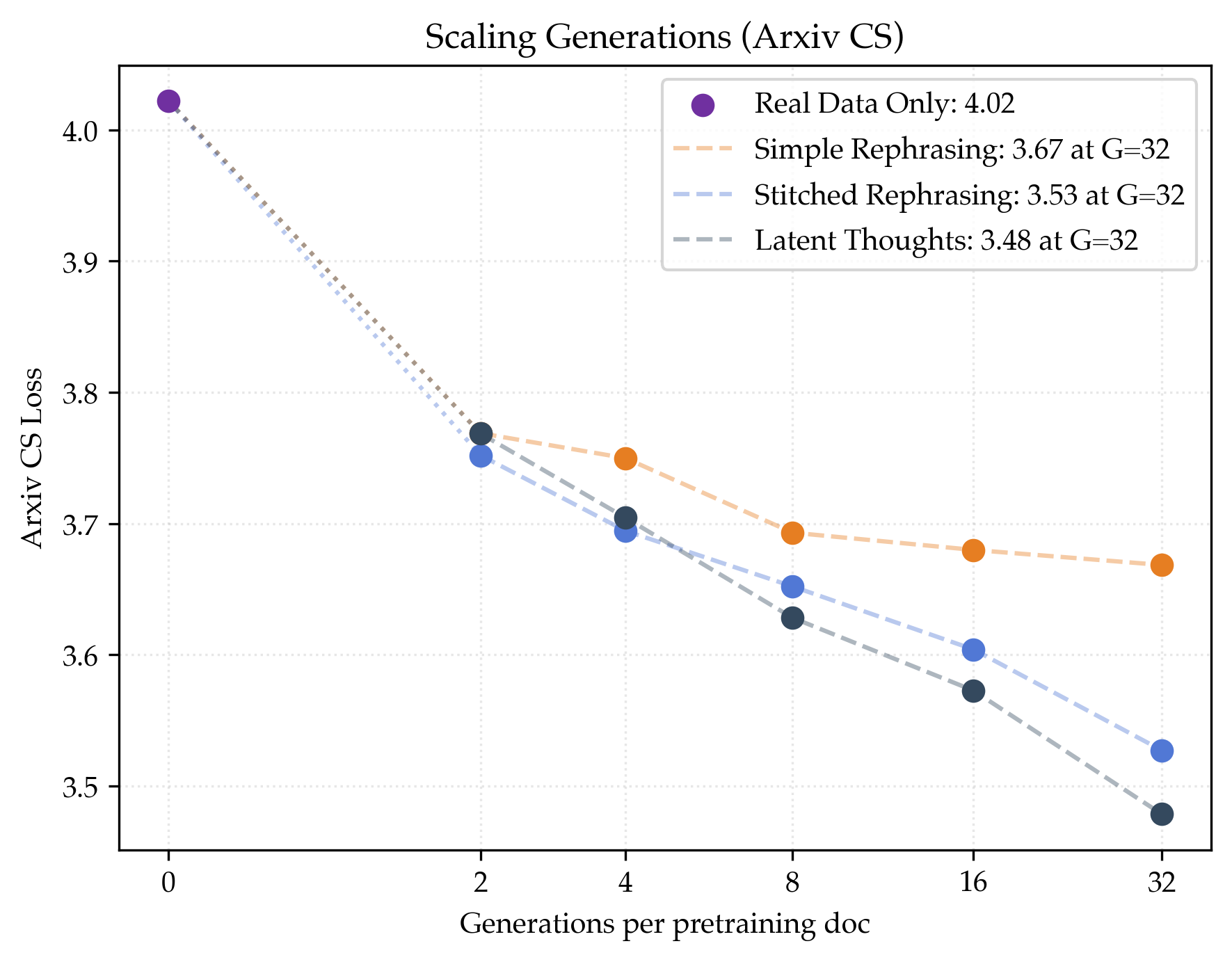}
    \end{minipage}

    \vspace{0.8em}

    \begin{minipage}[t]{0.48\textwidth}
        \centering
        \includegraphics[width=\linewidth]{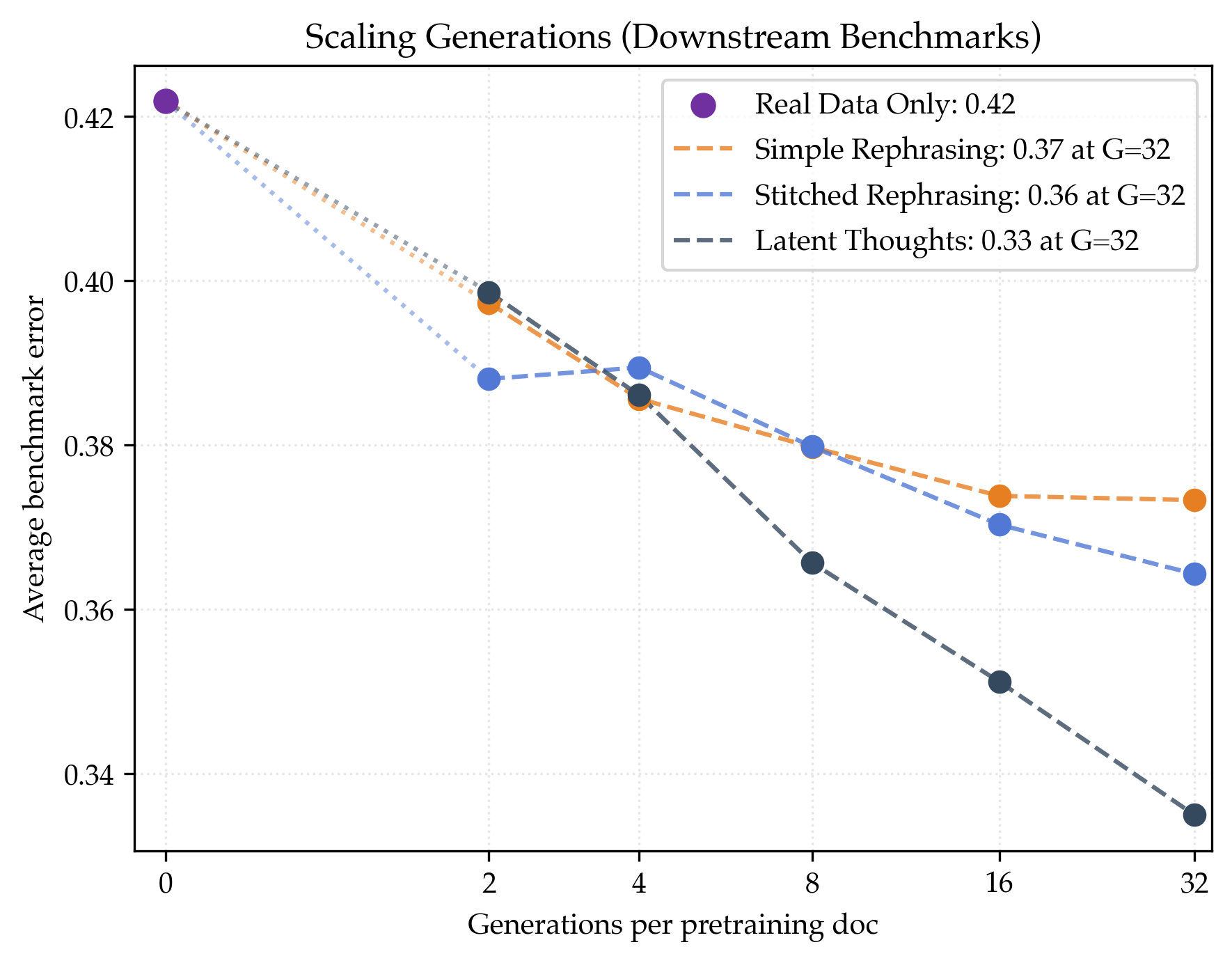}
    \end{minipage}
    \hfill
    \begin{minipage}[t]{0.48\textwidth}
        \vspace{-150pt}
        \caption{\textbf{Scaling generation count for stitching and latent thoughts.} We find that stitched rephrasing (blue points) and latent thoughts (gray points) improve i.i.d. loss, long-context loss, and downstream benchmark accuracy over simple rephrasing (orange points). The improvements scale in the number of generations per document and show less signs of plateauing.}
        \label{fig:sorting_gen_scaling}
    \end{minipage}

\end{figure}


\subsection{Scaling number of generations}\label{sec:scaling-gen-count-for-long}

\paragraph{Megadocs improve i.i.d. loss scaling and benchmark performance.} In Figure \ref{fig:sorting_gen_scaling}, we show how megadocs improve loss as we scale the number of generations per document. Stitched rephrasing and latent thoughts achieve $1.64\times$ and $1.80\times$ data efficiency, improving over simple rephrasing's $1.48\times$ data efficiency. Furthermore, the benefit widens as the number of generations increases: for example, the best loss improvement increases from $0.02$ at 4 generations to $0.07$ at 32 generations. The loss gains are also matched by downstream benchmarks, improving average accuracy by $6\%$ and $9\%$ (where simple rephrasing helps by $5\%$).

\paragraph{Megadocs improve long-context (and short-context).} Stitched rephrasing and latent thoughts especially improves scaling for long-context loss. When measuring loss on arXiv CS papers\footnote{Which are naturally long, like this one.} (Figure \ref{fig:sorting_gen_scaling} right), we observe a $0.14$ and $0.19$ loss improvement at $32$ generations (additional validation sets in Appendix \ref{app:additional-evals}). This is likely because the learning tasks induced by scaling document length promote capabilities useful for long-context tasks. However, we emphasize that megadocs improve loss on short context documents as well. In Appendix \ref{app:short-vs-long}, we find that megadocs still scale better than simple rephrasing on documents shorter than $600$ tokens.

We note that all hyperparameters are selected purely via locally optimal i.i.d.~loss, making long-context loss and benchmarks truly held-out evaluations. This may explain why the scaling looks less well-behaved for both metrics.

\subsection{Why do megadocs have better scaling?}

Our experiments show that megadocs result in better loss scaling as generation count grows. This is especially striking since the length of a megadoc quickly exceeds the model's context length of 4096 tokens. We hypothesize that the benefit comes from two different effects: a constant loss improvement from using megadocs combined with better scaling from enabling longer training.

\begin{figure}
    \centering
    \includegraphics[width=0.45\textwidth]{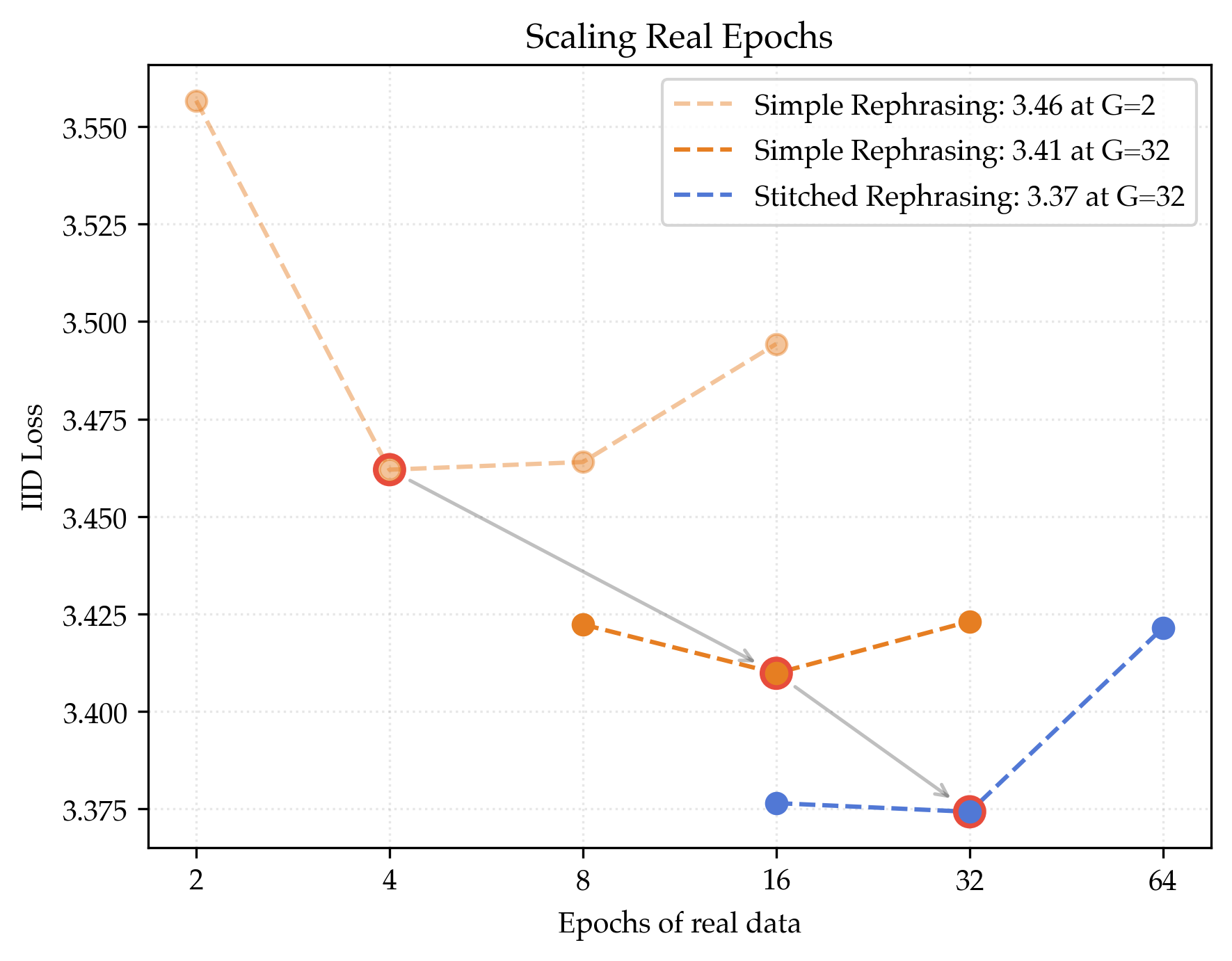}
    \includegraphics[width=0.45\textwidth]{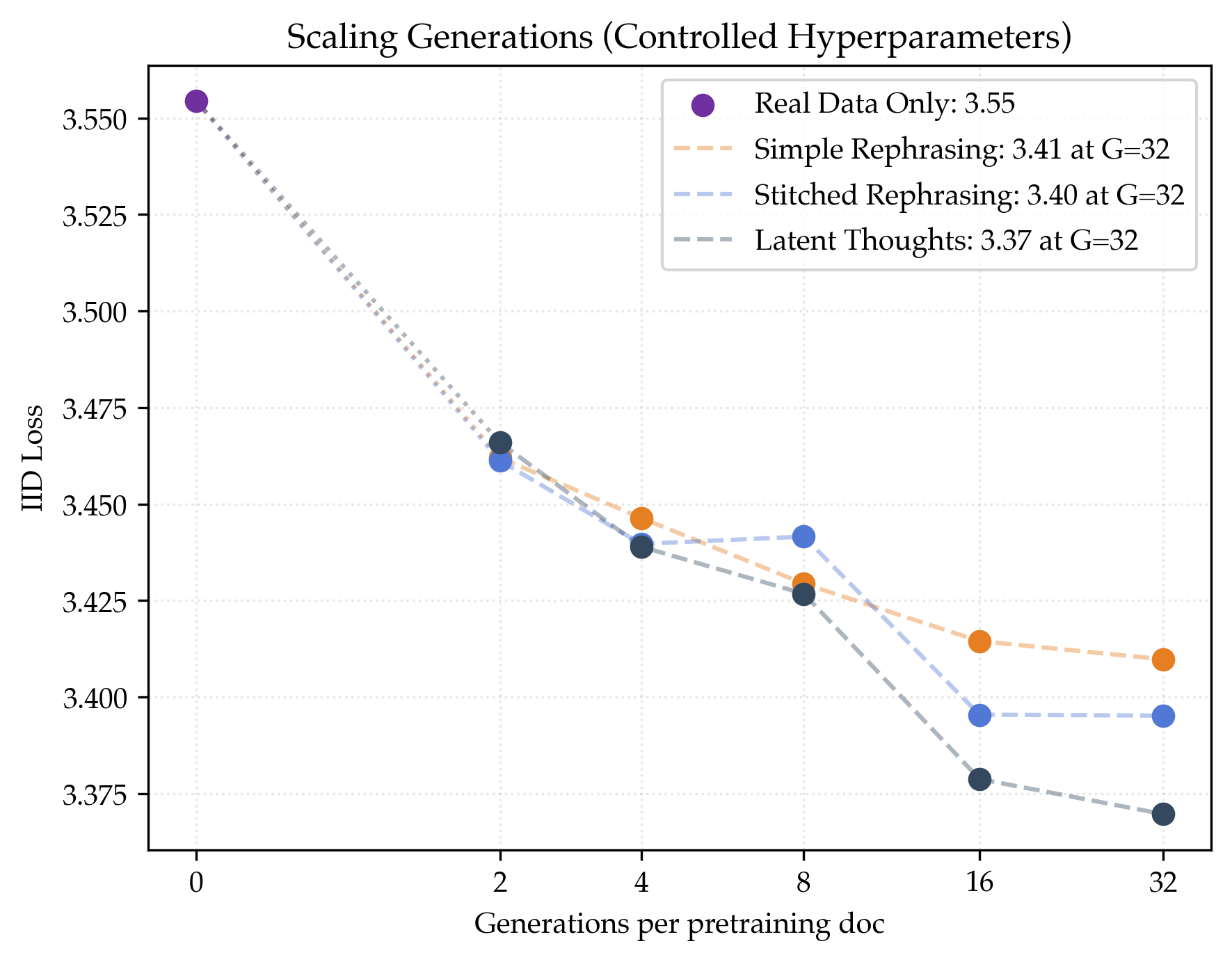}
    \caption{\textbf{Megadocs benefit from more steps on real and synthetic data.} We attribute the favorable scaling of megadocs to different optimal hyperparameters enabling longer training. For different synthetic data pools, we plot the optimal loss at each real data epoch count (left). We find that more synthetic tokens (higher $G$) allows for training for more real epochs, and megadocs push this effect further. Controlling for longer training by using the best simple rephrasing hyperparameters (right) shows that megadocs still improve but the loss win doesn't scale with $G$. The scaling is less clean in the right plot since the hyperparameters are tuned for simple rephrasing.}
    \label{fig:more_real_epochs}
\end{figure}

\paragraph{Megadocs enable longer training.} We hypothesize that megadocs improve scaling because their hyperparameters enable more training steps than simple rephrasing, extracting more value from the limited real and synthetic data. Importantly, megadocs such as stitched rephrases allow us to epoch the real data more before overfitting (16 to 32 epochs, Figure \ref{fig:more_real_epochs}, left) and and allow us to train at higher synthetic mixing fractions (0.75 to 0.9, Appendix \ref{app:optimal-hypers}). Together, the best hyperparameters for stitched rephrasing and latent thoughts are able to take $5\times$ more steps without overfitting. We can control for the effects of longer training by using the optimal simple rephrasing hyperparameters for the megadoc methods. Under this setting (Figure \ref{fig:more_real_epochs}, right), the favorable scaling of megadocs disappears and they provide a constant loss improvement at larger generation counts.



\section{Ensembling synthetically pre-trained models}

We test whether the benefits of our synthetic data algorithms compose with ensembling, one of the most effective data efficiency recipes shown to be central in \cite{kim2025pre}. When ensembling, we average the logits of models trained with different data orders and random initializations and find that loss decreases monotonically in ensemble member count with scaling exponent $1$.

\begin{figure}[h]
    \centering
    \includegraphics[width=0.8\textwidth]{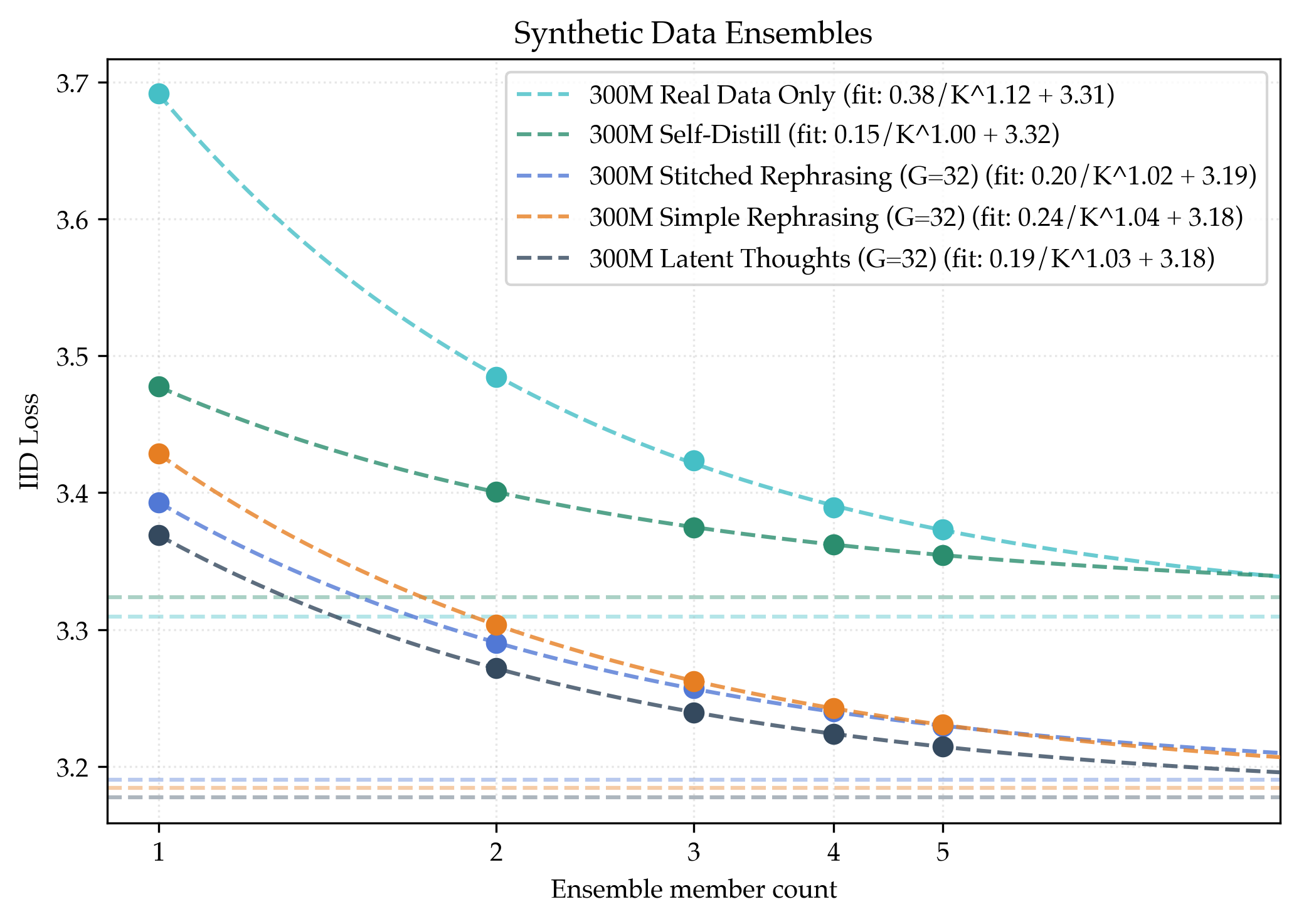}
    \caption{\textbf{Rephrasing composes with ensembling.} We measure the loss of ensembling synthetically pre-trained models as we increase the member count.
    For a given synthetic data recipe, we use the asymptote (horizontal dotted line) of a power law to estimate the loss as the member count approaches infinity. 
    Pre-training with self-distillation does not compose with ensembling as ensembling self-distilled models does not improve the asymptote. Meanwhile, simple rephrasing, stitched rephrasing, and latent thoughts all compose with ensembling and improve the asymptote as member count approaches infinity. Note that unlike simple rephrasing, the hyperparameters for stitched rephrasing and latent thoughts have not been tuned to improve the asymptote.}
    \label{fig:shuffled_ensemble}
\end{figure}

\paragraph{Self-distillation does not compose with ensembling} We first find that not all types of synthetic data compose with ensembling. \cite{kim2025pre} shows how self-distillation, an alternative form of synthetic data generated unconditionally from a previously trained model of the same architecture and data, can help improve the loss of a single model. In Figure \ref{fig:shuffled_ensemble}, we find that an ensemble of self-distilled models, though an improvement at finite member count, has the same asymptote (3.32) as an ensemble of standard 300M models (3.31). 
The lack of composition supports the theory in \cite{allenzhu2023understandingensembleknowledgedistillation} that self-distillation is effectively distilling a 2-ensemble into a single model.

\paragraph{Synthetic data augmentation does compose with ensembling} 
On the other hand, we find that simple rephrasing, stitched rephrasing, and latent thoughts all lower the ensemble member count asymptote relative to the standard pre-training ensemble asymptote. Each method improves over the ensembling asymptote for real data only models by at least $0.12$ loss. These synthetic data algorithms composing with ensembling suggests that their benefit is qualitatively distinct from ensembling and self-distillation. We note that unlike simple rephrasing, the hyperparameters for stitched rephrasing and latent thoughts have not been tuned to improve the asymptote and they have potential to perform better (Appendix \ref{app:ensemble-hypers}).



\section{Related Work}

\paragraph{Data efficiency.} Historically, many machine learning benchmarks were data-constrained \citep{726791, marcus-etal-1993-building, warstadt2023papersbabylmchallenge}, shaping a lot of the methods in the field. However, modern pre-training has been compute-constrained since its infancy, motivating compute-efficient algorithms \citep{hoffmann2022trainingcomputeoptimallargelanguage,kaplan2020scalinglawsneurallanguage}. Due to massive compute scaling, there has been a returning interest to data-constrained pre-training algorithms with proposed methods including epoching \citep{muennighoff2023scalingdataconstrainedlanguagemodels}, diffusion language models \citep{prabhudesai2025diffusionbeatsautoregressivedataconstrained}, and energy-based models \citep{gladstone2025energybasedtransformersscalablelearners}. Our paper measures the effectiveness of synthetic data under the formalization of data-efficient pre-training laid out in \cite{kim2025pre} with the practical deviation of an external synthetic data generator. 

\paragraph{Synthetic data.} Synthetic data has been critical for distillation \citep{eldan2023tinystoriessmalllanguagemodels}, verifiable domains \citep{zelikman2022starbootstrappingreasoningreasoning}, and alignment \citep{bai2022constitutionalaiharmlessnessai}. Data augmentation via synthetic model generations is recently finding more popularity in vision \citep{azizi2023syntheticdatadiffusionmodels,trabucco2025effectivedataaugmentationdiffusion} and language to improve data quality and handle data constraints, used in many model/data releases \citep{kimiteam2025kimik2openagentic,su2025nemotroncctransformingcommoncrawl,olmo2025olmo3,benallal2024cosmopedia}. The predominant form of synthetic data is rephrasing \citep{maini2024rephrasingwebrecipecompute,nguyen2025recyclingwebmethodenhance,datologyai2025beyondweblessonsscalingsynthetic,kang2025demystifyingsyntheticdatallm,qin2025scalinglawssyntheticdata,allenzhu2024physicslanguagemodels31,niklaus2026_the_synthetic_data_playbook_generating_trillions_of_the_finest_tokens} with other alternatives emerging \citep{yang2024syntheticcontinuedpretraining,ruan2025reasoninglearnlatentthoughts,lin2025learningfactsscaleactive,yang2025syntheticbootstrappedpretraining,qin2025scalinglawssyntheticdata}. Among these works, \cite{yang2025syntheticbootstrappedpretraining,kang2025demystifyingsyntheticdatallm,maini2024rephrasingwebrecipecompute} also find that synthetic data can improve validation loss. Our work distinguishes itself from prior work by measuring scaling in loss under the number of synthetic generations, introducing the idea of megadocs, and composing methods with over-parameterization, epoching, regularization, and ensembling.


\paragraph{Relationships across documents.} Some pre-training algorithms have found benefits to modeling the relationship between various pre-training documents. Synthetic Bootstrap Pre-training (SBP) found a benefit to modeling the relationships between pre-training documents that are close in embedding distance \citep{yang2025syntheticbootstrappedpretraining}. In-context Pre-training (ICPT) found a benefit by sorting documents that are close in embedding distance to be next to each other  \citep{shi2024incontextpretraininglanguagemodeling}. In fact, ICPT can be interpreted as a multi-document generalization of the SBP synthesizer training and we speculatively believe mixing may play the same role as distillation back into a student for i.i.d.~loss. This would suggest that stitched rephrasing can get the benefits of rephrasing, SBP, and ICPT simultaneously. We detail this potential relationship in more detail in Appendix \ref{app:sbp-icpt}.

\section{Discussion}

\subsection{Using a stronger generator}\label{sec:stronger-generator}

In this paper, since it is too expensive to pre-train capable generators from scratch, we use an external generator that is more capable than the student following prior work \citep{maini2024rephrasingwebrecipecompute,yang2024syntheticcontinuedpretraining,benallal2024cosmopedia}. Nonetheless, there are at least three reasons to be optimistic that our synthetic data improvements are not purely artifacts of distillation and would be strong, if not stronger, in a data/compute regime where we could pre-train generators from scratch. 
\begin{enumerate}
    \item \textbf{Synthetic data helps more for larger students (Appendix \ref{app:student-scaling-ablation}).} The primary concern is that rephrasing helps the student only because it is significantly weaker than the generator. To address this concern, we measure the loss improvement of stitched rephrasing and latent thoughts after scaling up the student model $5\times$ for the same generator. We find that for all three synthetic data algorithms, the loss improvements are actually larger when using the stronger student despite the baseline loss itself being lower. This is consistent with prior work on student/generator scaling for synthetic data augmentation as discussed in the next point \citep{kotha2025selfimprovement}.
    \item \textbf{Self-improvement via synthetic data (Appendix \ref{app:student-generator-scaling}).} It is commonly argued that if the generator comes from the same data as the student, synthetic data can not help due to the data processing inequality. However, prior work shows examples where simple rephrasing and latent thoughts admit self-improvement.
    \begin{enumerate}
        \item For rephrasing, \cite{kotha2025selfimprovement} shows that Llama 3.1 8B Instruct can improve itself on a data-constrained factuality task simply by training on self-generated rephrases. Furthermore, the student scaling in that work agrees with our ablations that synthetic data from a fixed teacher helps more for more capable students, not less.
        \item For inter-document modeling, \cite{yang2025syntheticbootstrappedpretraining} shows how modeling the relationship between related documents can improve a model trained on the same original data. This gives reason to be optimistic about the benefit of modeling inter-document relationships via megadocs in the self-improvement setting.
        \item For latent thoughts,  \cite{zelikman2024quietstarlanguagemodelsteach,ruan2025reasoninglearnlatentthoughts,hatamizadeh2026rlpreinforcementpretrainingobjective} show that generating latent synthetic data from the student model in an on-policy fashion provides a useful signal that improves over the course of training. The latent thoughts augmentation in this paper is an off-policy version of these methods where the thought generator is frozen and not learned. 
    \end{enumerate}
    \item \textbf{Data augmentation in vision.} In vision, simple data augmentations such as blurring an image are well-established to help for almost every model/data scale \citep{steiner2022trainvitdataaugmentation}. It is possible that rephrasing operates similarly to such augmentations after being unlocked given sufficient compute. Multiple reports corroborate the view that rephrasing is a simple capability since scaling the rephraser stops helping past a certain scale \citep{datologyai2025beyondweblessonsscalingsynthetic, lin2025learningfactsscaleactive, maini2024rephrasingwebrecipecompute, kang2025demystifyingsyntheticdatallm,niklaus2026_the_synthetic_data_playbook_generating_trillions_of_the_finest_tokens,kotha2025selfimprovement}.
\end{enumerate}

Nonetheless, there are many practical applications of synthetic data utilizing a stronger generator such as pre-training open-source models \citep{olmo2025olmo3, benallal2024cosmopedia} and data-constrained continued pre-training \citep{kotha2026replayingpretrainingdataimproves, yang2024syntheticcontinuedpretraining}. For such regimes, measuring i.i.d.~loss, scaling generation count, and utilizing megadocuments should similarly work.


\section{Acknowledgments}

We thank \texttt{us-central2}, Yangjun Ruan, Kaiyue Wen, Nathan Hu, Seungju Han, Steven Cao, Mike Lewis, and members of \href{https://yejinc.github.io/}{xlab} for their helpful discussions and feedback. This work is a part of the \href{marin.community}{Marin Project} and the compute is supported by the Google TPU Research Cloud (TRC). 
KK was supported by the National Science Foundation Graduate
Research Fellowship Program under DGE-2146755. Any opinions,
findings, and conclusions or recommendations expressed in this material are those of the
author(s) and do not necessarily reflect the views of the National Science Foundation.
NH was supported under NSF \#2302701 and the HAI Hoffman-Yee grant program. YC was supported by the AI Research Hub Project through KAIST and Singapore DSO. TH was supported by a grant by HAI, DSO labs, gifts from Open Philanthropy, Amazon, Schmidt Sciences, the Tianqiao and Chrissy Chen Foundation and a grant under the NSF CAREER IIS-2338866, ONR N00014-24-1-2609, and DARPA Cooperative Agreement HR00112520013. PL was supported by DARPA Cooperative Agreement HR00112520013. This work does not necessarily reflect the position or policy of the government and no official endorsement should be inferred.

\newpage
\bibliographystyle{abbrvnat}
\bibliography{references}


\newpage
\appendix
\tableofcontents

\newpage 

\section{Using a stronger generator}\label{app:stronger-generator}

We offer a more detailed explanation of the arguments in Section \ref{sec:stronger-generator} that synthetic data is likely to help even when the generator is derived from the same data given sufficient compute, specifically for rephrasing.

\subsection{Student scaling}\label{app:student-scaling-ablation}

\begin{figure}[ht]
\centering
\includegraphics[width=0.8\textwidth]{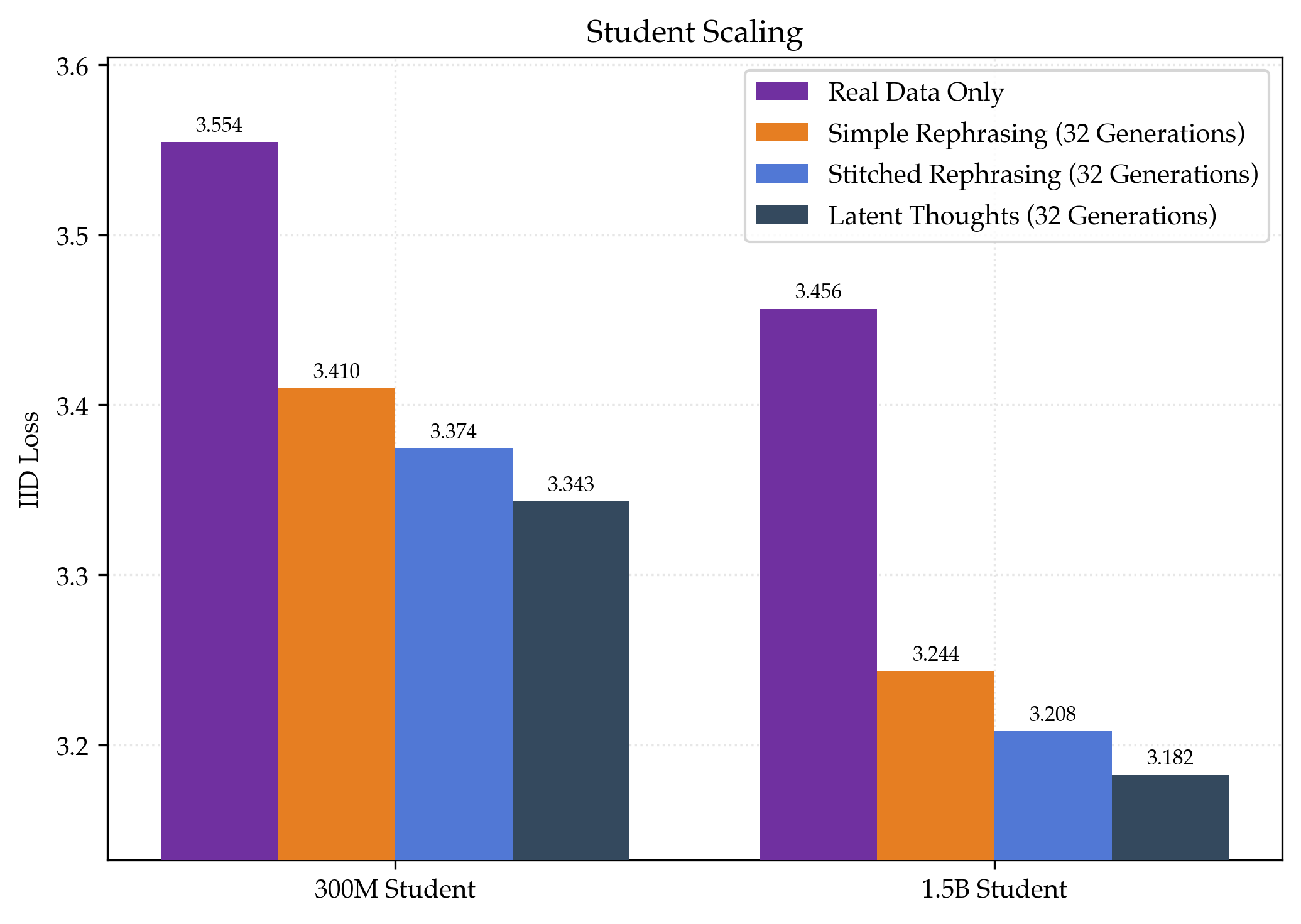}
\vspace{0.6em}
\resizebox{0.6\textwidth}{!}{
\begin{tabular}{lrrr}
\toprule
\textbf{Method} & \textbf{300M Win} & \textbf{1.5B Win} & \textbf{$\Delta$}\\
\midrule
Simple Rephrasing & 0.144 & 0.212 & +0.068 \\
Stitched Rephrasing & 0.18 & 0.248 & +0.068 \\
Latent Thoughts & 0.211 & 0.274 & +0.063 \\
\bottomrule
\end{tabular}}
\caption{\textbf{Student scaling}. We pre-train 1.5B student models using the same set of synthetic data methods. At each scale, we measure the loss win of each method over the no synthetic data baseline (purple). We observe a larger loss win at the 1.5B scale than the 300M scale for all methods.}
\label{fig:student_scaling}
\end{figure}

We ablate the effect of having a stronger student by pre-training 1.5B models under the same setup: 200M tokens of DCLM data and synthetic data generated by Llama 3.1 8B instruct. See Appendix \ref{app:training-details} for details on architecture. When we compare the loss win of our synthetic data methods over the real data only baseline, we find that the loss win scales with student size (Figure \ref{fig:student_scaling}). For all three methods, the $\Delta$ between the loss win at 1.5B scale and 300M scale is at least 0.06. Furthermore, our 1.5B runs are not locally optimal due to compute constraints and were only selected over a fixed grid of $\text{weight decay} \in \{0.4,0.8,1.6,3.2\},\text{real epochs} \in \{4, 8\}, \text{mixing fraction} \in \{0.5, 0.75\}$. Since the 1.5B losses are upper bounds, it's likely that the true gap in student scaling is even larger.

\subsection{Self-improvement via synthetic data (rephrasing)}\label{app:student-generator-scaling}

\begin{figure}
    \centering
    \includegraphics[width=0.95\textwidth]{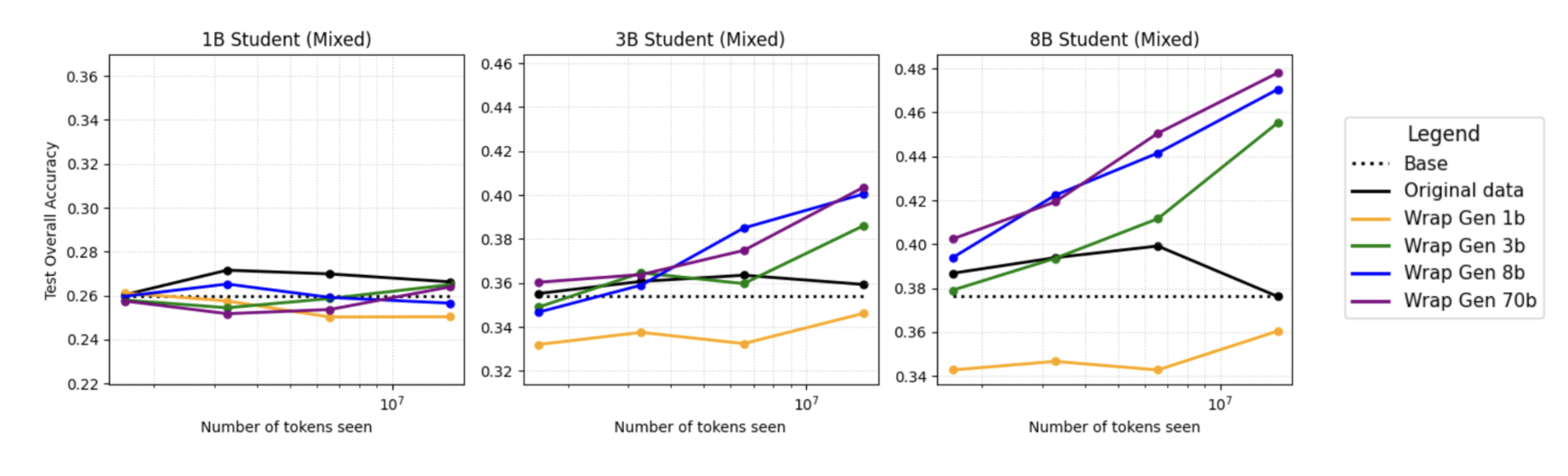}
    \caption{\textbf{Student and generator scaling laws for continued pre-training.} For a small set of 1.3M tokens, we measure QA accuracy when training utilizing Rephrase the Web \citep{maini2024rephrasingwebrecipecompute}. We find that self-improvement is possible (3B student and generator, 8B student and generator). Additionally, rephrasing only emerges with sufficient scale, stronger students benefit more from synthetic data, and synthetic data improvements eventually stop scaling in generator capability. Figure taken from preliminary experiments in \citep{kotha2025selfimprovement}.}
    \label{fig:student_generator_scaling}
\end{figure}

\paragraph{Self-improvement demonstration for rephrasing.} We first share a clear example of self-improvement via synthetic data augmentation from \cite{kotha2025selfimprovement}. This work studies data efficiency for Quality \citep{pang-etal-2022-quality}: 1.3M tokens of articles with an associated QA comprehension benchmark (setting introduced in \cite{yang2024syntheticcontinuedpretraining}). When performing continued pre-training with Llama 3.1 8B Instruct, epoching Quality gives up to a $3.5\%$ accuracy improvement. However, after generating 840M tokens of synthetic rephrases via the student model using Rephrase the Web \citep{maini2024rephrasingwebrecipecompute}, the model can improve itself by $16.6\%$. In early experiments with a different continued pre-training dataset, the authors also found that this improvement translated to i.i.d.~validation loss when mixed with real data. The experiments in \cite{kotha2025selfimprovement} additionally study how accuracy changes as a function of student and generator capabilities, main takeaways below.

\paragraph{Observation 1: Generators need to be sufficiently capable at rephrasing.} In Figure \ref{fig:student_generator_scaling}, we see that using a weak generator (Llama 1B Instruct, yellow lines) does not generate meaningful synthetic data. In fact, it can hurt at larger scales. This is why we are forced to use a larger synthetic data generator in our experiments instead of our models pre-trained from scratch on 200M tokens.

\paragraph{Observation 2: Stronger student benefits more from rephrasing.} When increasing the capability of the student (Llama 1B to Llama 8B, left to right), we find that stronger students benefit more from synthetic data than weaker students. Therefore, even though the experiments in this paper uses a low capability student, this does not imply the gains would be smaller for stronger students. The results in Appendix \ref{app:student-scaling-ablation} corroborate the finding of stronger students benefiting more from synthetic data.

\paragraph{Observation 3: Scaling the generator doesn't help past a certain point.} We find that making the generator stronger does not significantly help past a certain point (8B Instruct to 70B Instruct, blue to purple line). This phenomenon has been observed in many prior works \citep{datologyai2025beyondweblessonsscalingsynthetic, lin2025learningfactsscaleactive, maini2024rephrasingwebrecipecompute, kang2025demystifyingsyntheticdatallm,niklaus2026_the_synthetic_data_playbook_generating_trillions_of_the_finest_tokens}. Therefore, it is possible that we get to close to the upper limits of generator scaling via rephrasing simply by utilizing Llama 3.1 8B Instruct. This also contributes to our view as rephrasing as an augmentation rather than a self-improvement capability.

\paragraph{Implications for our work.} Suppose our goal was to study self-improvement for data-efficient pre-training on the internet scale data instead of 200M tokens. Since we do not have the compute to pre-train such models from scratch, we have to use a smaller student (Observation 1). However, we are optimistic that our synthetic data improvements would translate to the regime where students are stronger (Observation 2) and that the gap between the student and teacher isn't the key ingredient to the error reduction (Observation 3). It would be interesting for future work to test our findings at scale from scratch.

\section{Synthetic data hyperparameters}\label{app:synth-data-hypers}

\subsection{Choice of mixing parameterization and order of limits}\label{app:parameterization}

Our paper takes the stance of fixing a number of generations, training under unlimited compute (via epoching the real data and synthetic data), and then varying the generation compute budget. This particular order of limits is important for clean scaling for the following reasons.

\begin{enumerate}
    \item Ultimately, we want to characterize the performance of algorithms under infinite generation ($G \to \infty$) and training compute. Our parametrization estimates this with a double limit: allowing epoching of the synthetic stream first measures infinite train compute performance for fixed $G$, and then sending $G \to \infty$ measures the best loss under infinite generation compute.
    \item Always training on fresh synthetic data would naturally result in exactly 1 epoch on the synthetic data. Unfortunately, this is not expected to be monotonic in the number of generations since too much synthetic data could result in low representation of real data during training and overfitting to the synthetic distribution.
    \item A second natural choice is limiting up to 1 epoch of synthetic data. Our approach of fully tuning epoching on the synthetic data, as opposed to limiting training on it at most once, gives strictly lower loss at every choice of $G$. Our scaling gives a lower bound on the loss of unepoched synthetic data. 
\end{enumerate}

\subsection{Tuning hyperparameters}\label{app:optimal-hypers}

We are interested in finding the best setting of hyperparameters (e.g. learning rate, epoch count, weight decay, mixing fraction) given a fixed pool of real and synthetic data to train on. We discretize the space of hyperparameters and search for locally-optimal hyperparameters as done in \citep{kim2025pre}. Given a $d$-dimensional grid over hyperparameters, we define a point on this grid as locally optimal if deviating along any of the $d$ dimensions in either direction doesn't improve loss. 

\begin{figure}[h]
    \centering
    \includegraphics[width=0.45\textwidth]{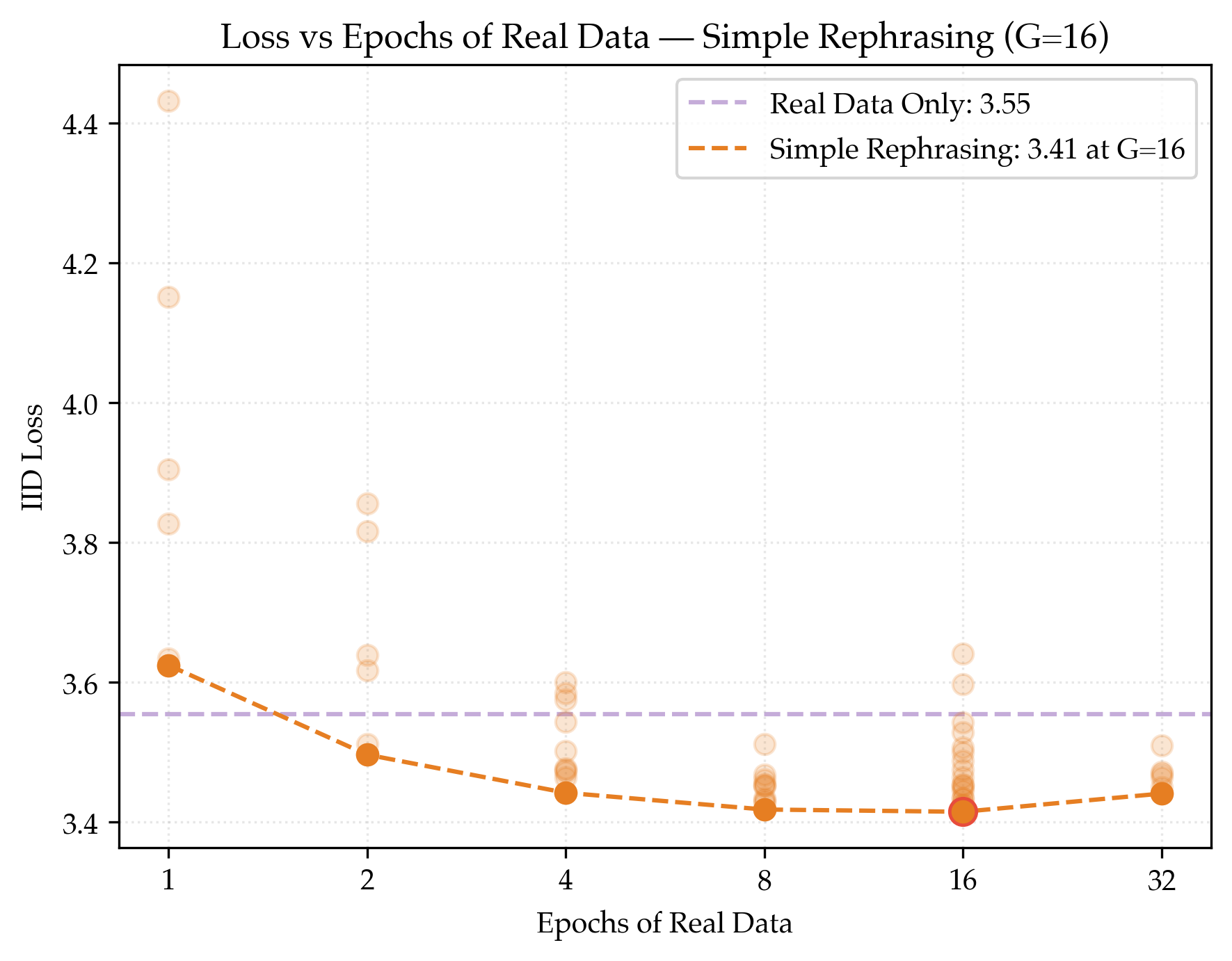}
    \includegraphics[width=0.45\textwidth]{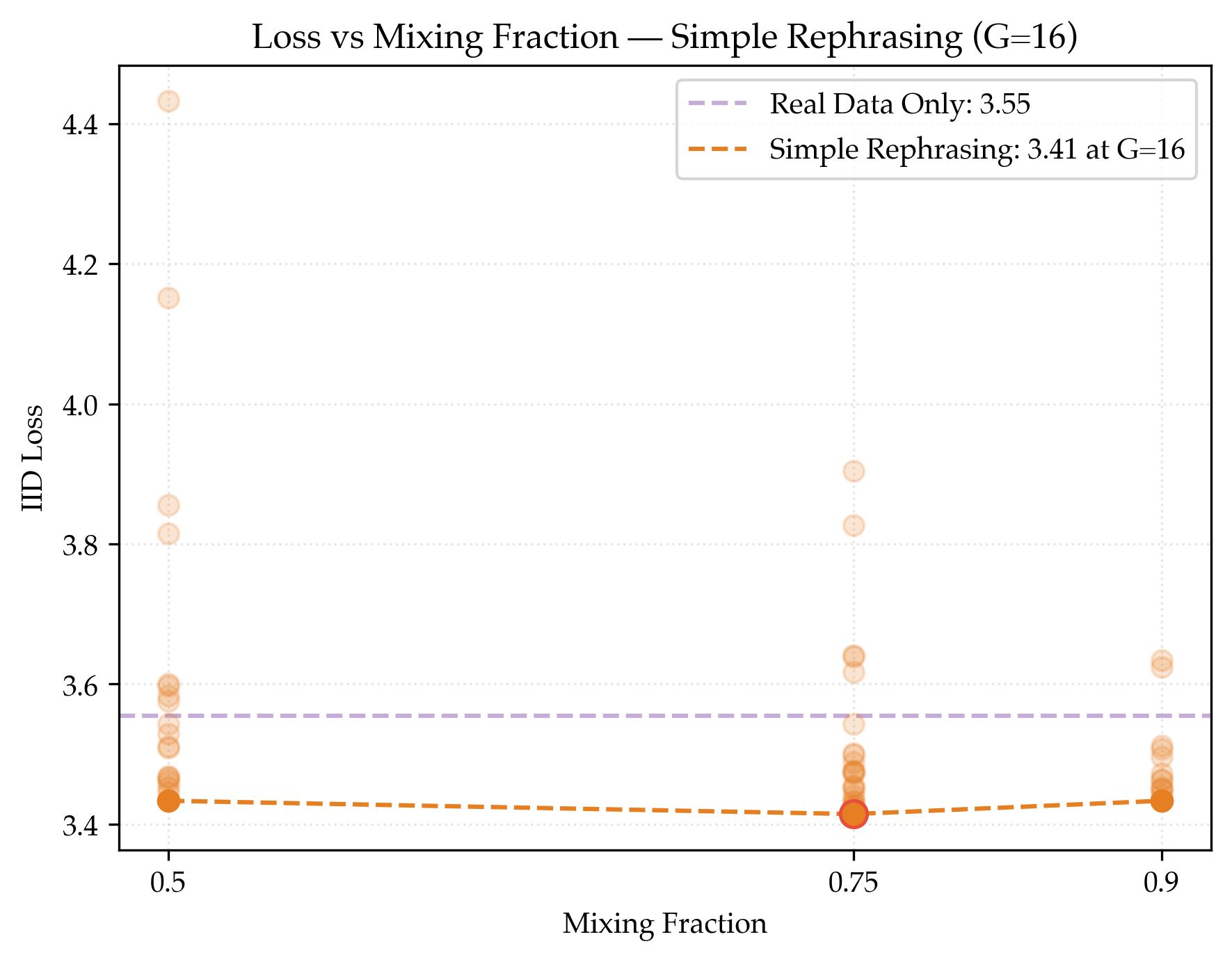}
    \caption{\textbf{Convexity in hyperparameters.} For a fixed pool of synthetic data (16 rephrases per document), we show how the optimal loss changes as we vary two hyperparameters: real data epochs and mixing fraction. For our synthetic data scaling experiments, we pick the "locally optimal" solution so that deviating from any of the hyperparameters doesn't improve loss.}
    \label{fig:convex_hps}
\end{figure}

To search for locally optimal solutions, we first 
seed the search with initial runs around a best guess for optimal hyperparameters. We then iteratively take the best run, run all of its neighbors, and repeat if one of the neighbors is better. Otherwise, we call the run ``certified'' and terminate the search. In Figure \ref{fig:convex_hps}, we show an example of this convex certification for two hyperparameters. 

\begin{figure}
    \centering

    \begin{minipage}[t]{0.48\textwidth}
        \centering
        \includegraphics[width=\linewidth]{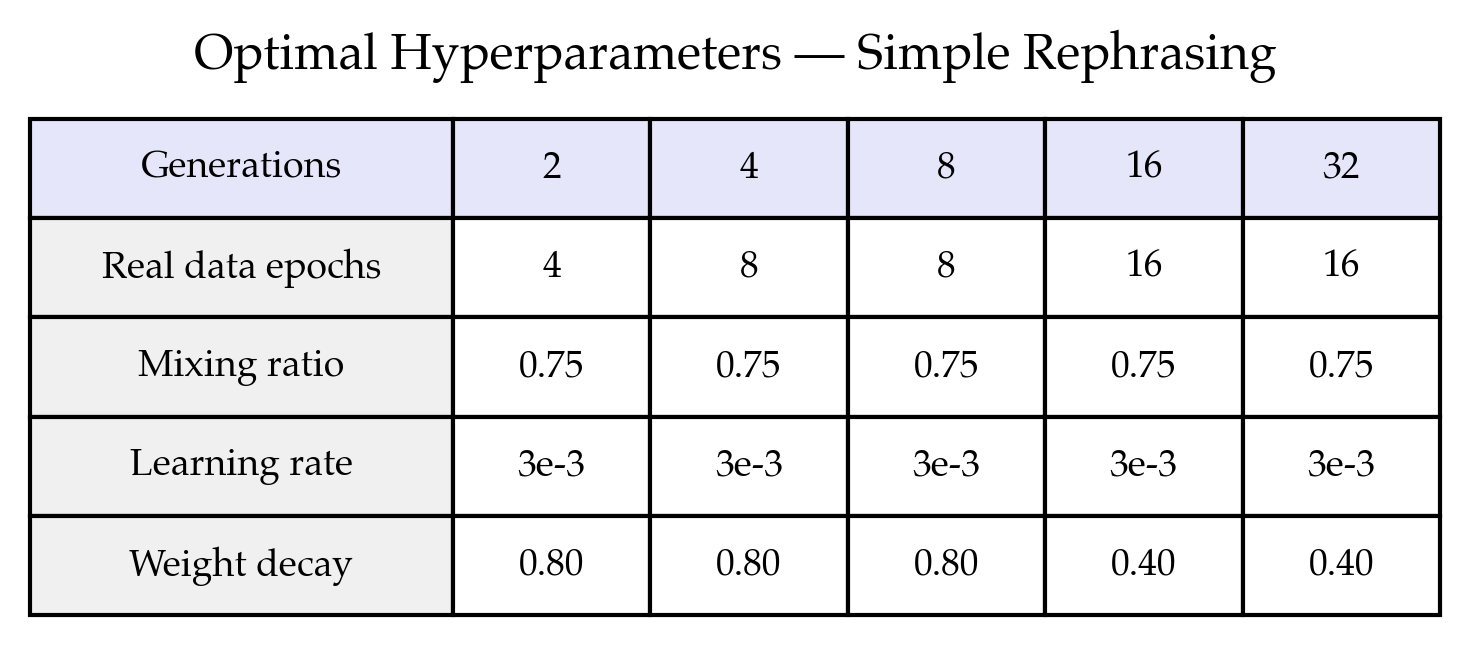}
    \end{minipage}
    \hfill
    \begin{minipage}[t]{0.48\textwidth}
        \centering
        \includegraphics[width=\linewidth]{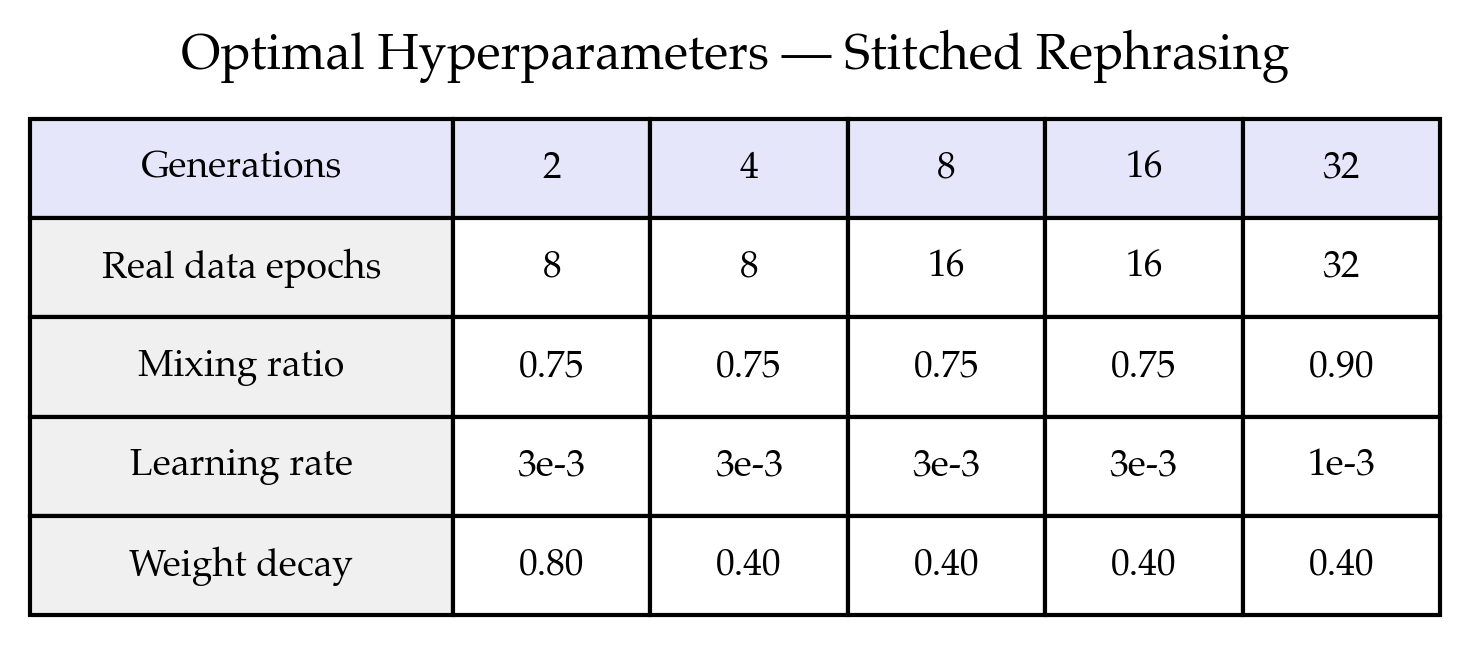}
    \end{minipage}

    \vspace{0.8em}

    \begin{minipage}[t]{0.48\textwidth}
        \centering
        \includegraphics[width=\linewidth]{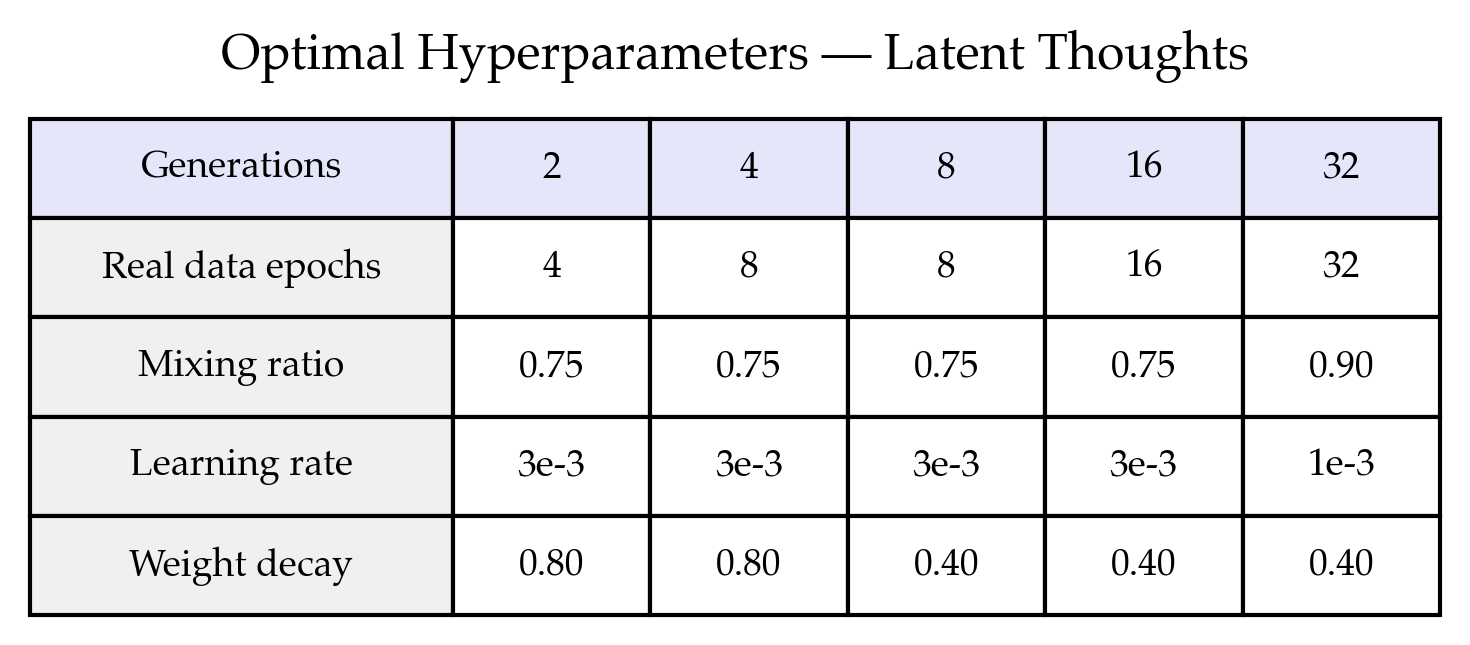}
    \end{minipage}
    \hfill
    \begin{minipage}[t]{0.48\textwidth}
        \vspace{-90pt}
        \caption{\textbf{Optimal hyperparameters.} We visualize the locally optimal hyperparameters for each of the synthetic data methods we scale in our paper. We observe that megadoc-based methods can epoch the real data more and sustain a higher mixing ratio without overfitting, particularly at the highest synthetic generation counts.}
        \label{fig:optimal_hypers}
    \end{minipage}

\end{figure}

Our specific discretization considers learning rates in $\{0.001, 0.003, 0.01\}$, weight decay in $\{0.1, 0.4, 0.8, 1.6\}$, epoch count in $\{1,2,4,8,16,32\}$, and mixing fraction in $\{0.5,0.75,0.9\}$. Due to compute constraints, we set bounds on the search space, so we set a maximum epoch count of $32$ and a maximum mixing fraction of $0.9$. Only 2 of our runs trigger these maximum bounds: stitched rephrasing and latent thoughts at 32 generations per pretraining doc. For each synthetic data method, we visualize the optimal hyperparameters in Figure \ref{fig:optimal_hypers}.  

\subsection{Ensemble hyperparameters}\label{app:ensemble-hypers}

Due to compute constraints, we don't search for locally optimal hyperparameters for our synthetic data ensembles as prior work \citep{kim2025pre} observed that the optimal hyperparameters for the lowest ensemble asymptote were different from the optimal hyperparameters for the best single model loss. Instead, \cite{kim2025pre} used the heuristic of doubling the epoch count and halving the weight decay of the best single model hyperparameters. We adopt a version of this heuristic by taking the best synthetic data hyperparameters and only doubling the epoch count.   

When composing our megadoc methods with ensembles, we opt to use the hyperparameters of the simple rephrasing ensemble for a more fair comparison. It's likely that using the above heuristic on the megadoc hyperparameters would further improve the ensemble asymptote.  

\section{Cross-document attention}

\begin{figure}
    \centering
    \includegraphics[width=0.8\textwidth]{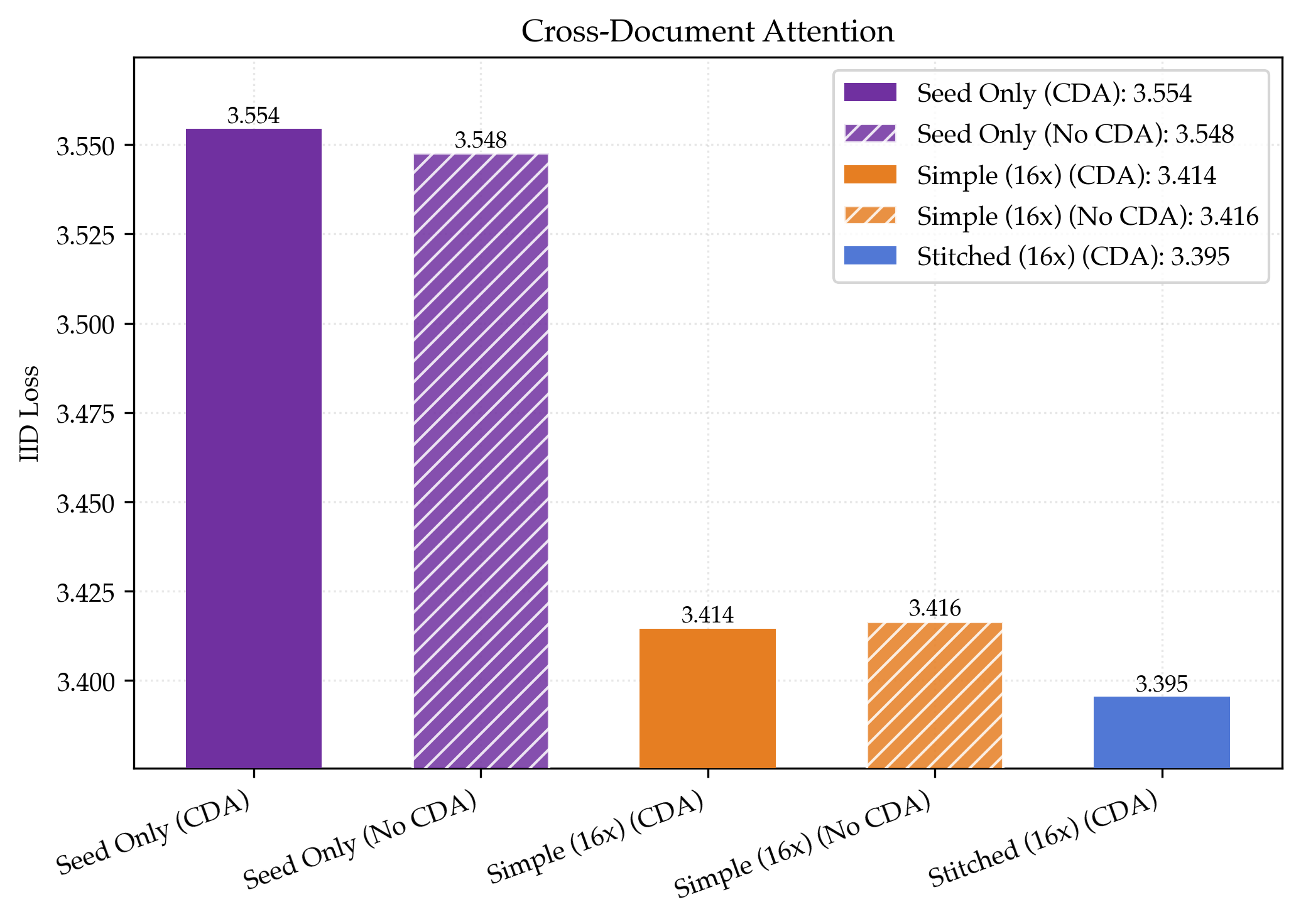}
    \caption{\textbf{Cross-document attention ablation.} In the data-eficient regime with only real data, it is better to mask attention between separate documents in the same context window (purple bars). When also training on rephrased data, the answer changes (orange bars). The benefit of stitching additionally justifies disabling cross-document attention masking.}
    \label{fig:cda_ablations}
\end{figure}

Prior work has shown that in a compute-efficient training regime, masking attention between documents in the same context window results in better loss, presumably due to decreasing distractor tokens \citep{Zhao_2024}. We reproduce this finding in the data-efficient regime, finding that it is best to enable cross-document masking for the baseline (Figure \ref{fig:cda_ablations}, purple bars).

 We then ablate on enabling cross-document masking for synthetic data. We find the effect of masking to be within noise for simple rephrasing (orange bars). As stitching requires no masking, it is worth disabling cross-document attention during training to support stitching (blue bar).

 In general, we find enabling cross-document attention to help for related data. Stitching enables this for synthetic data, but in earlier experiments during this project conducted on natural data with more ``related'' documents (MultiNews \citep{fabbri2019multinewslargescalemultidocumentsummarization} and a contiguous subsequence of crawl time sorted DCLM), we also found a large benefit to disabling masking (only when epoching).



\section{Stitching}

\subsection{Mixing with stitching improves i.i.d.~loss}\label{app:mixing-iid}

\begin{figure}
    \centering
    \includegraphics[width=0.8\textwidth]{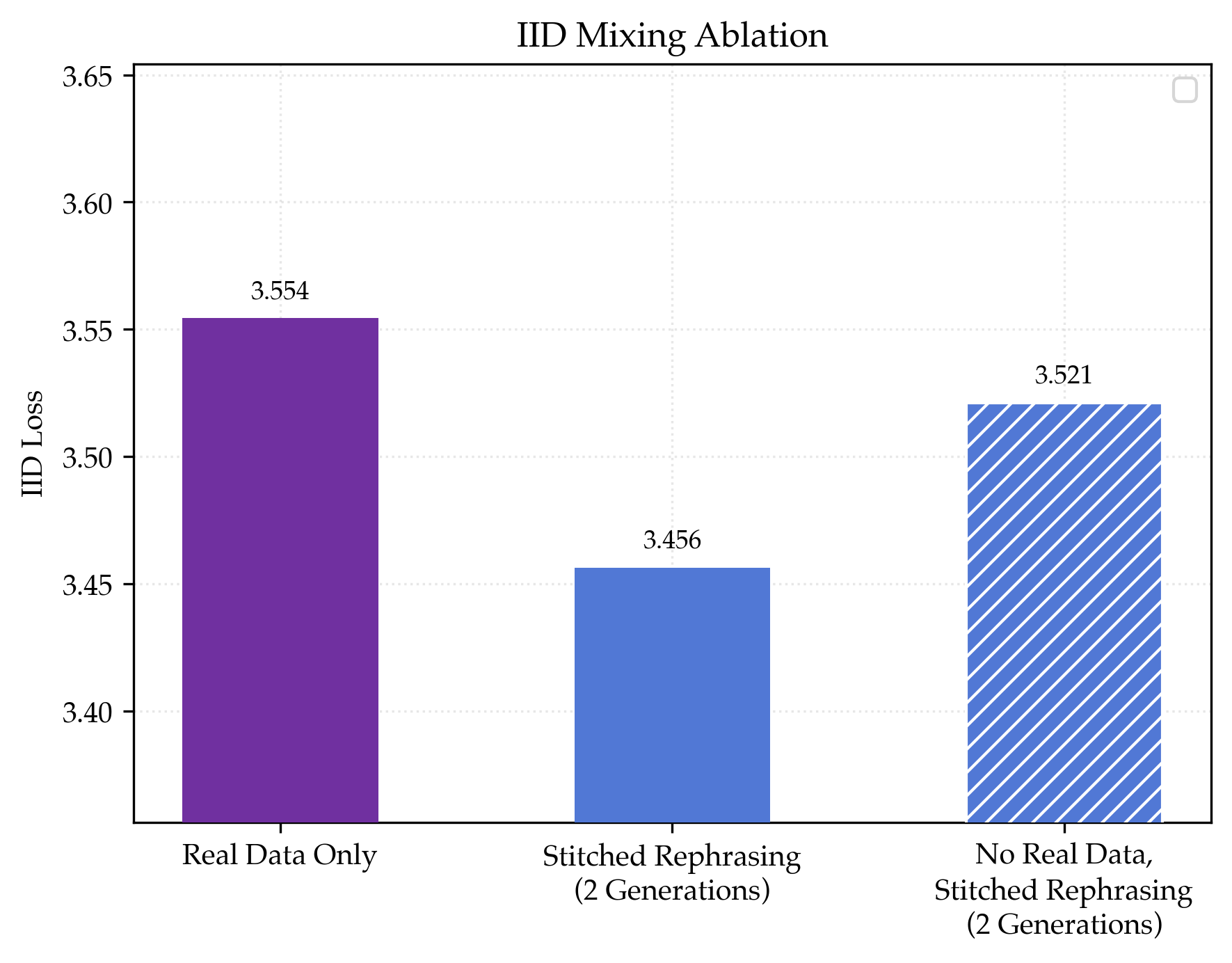}
    \caption{\textbf{Mixing ablation.} We briefly ablate the importance of mixing in real data. Given a small synthetic data pool (2 rephrases per document), we find that also training on the real data vs. not training on it substantially improves loss from 3.521 to 3.456. Consequently, we always mix real data and search for a locally optimal mixing fraction.}
    \label{fig:iid_mixing_loss}
\end{figure}

Prior work on in-context pre-training has primarily shown its benefit on long-context loss and benchmarks \citep{shi2024incontextpretraininglanguagemodeling}, while more recent work on synthetic data has shown the importance of mixing in real data for i.i.d~loss \citep{kang2025demystifyingsyntheticdatallm}. 
In Figure \ref{fig:iid_mixing_loss}, we ablate the effect of mixing real data on stitched rephrasing. Given a small synthetic data pool generated with 2 rephrases per document, we lightly search for the best ``no real data, stitched rephrasing'' baseline by tuning the number of synthetic epochs $\{4,8\}$ and the weight decay $\{0.4,0.8,1.6\}$. We find that 70\% of the loss improvement of stitched rephrasing comes from mixing in real data, so we always mix real data and search for locally optimal mixing fractions in our scaling experiments.

\subsection{Real data in synthetic stream helps}\label{app:real-in-synthetic}

\begin{figure}
    \centering
    \includegraphics[width=0.8\textwidth]{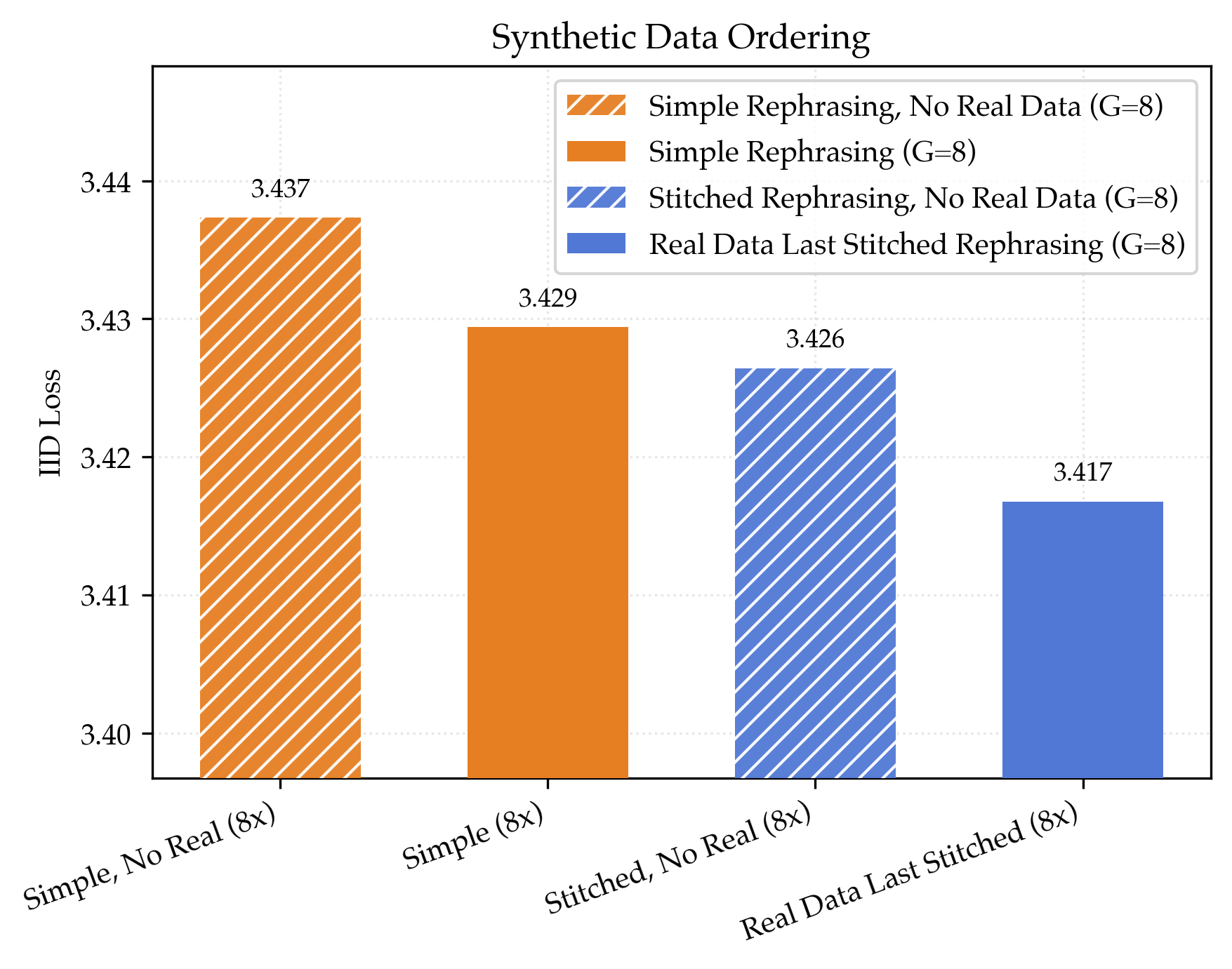}
    \caption{\textbf{Including real document in synthetic stream.} We show how stitching is a benefit over simple rephrasing both when there is no real data in the synthetic stream (striped blue vs striped orange) and when there is real data (solid blue vs solid orange). Since including the real document in the synthetic stream is helpful for both simple rephrasing (solid orange vs striped orange) and stitched rephrasing (solid blue vs striped blue), we opt to include a real document in the synthetic stream for all experiments in the main body.}
    \label{fig:sorting_nodoc_ablations}
\end{figure}

Surprisingly, when training on synthetic data, we find that including the real web document in the synthetic stream always improves loss. In Figure \ref{fig:sorting_nodoc_ablations}, regardless of whether we train on simple rephrases (orange) or stitched rephrases (blue), including the real document (solid vs. striped) improves loss. Since keeping the real document always helps, we use this for all synthetic data experiments in the main body. 

The benefit of keeping the real document is larger when training on stitched rephrases, so we ran additional ablations on where the real document should go in Figure \ref{fig:sorting_ablations} but only lightly explored the design space. Other options could include keeping it at the end of each context window, but it's likely that having too much real data in the synthetic stream can cause overfitting effects. More broadly, we believe there's a large design space for megadocs such as changing the prompting strategy to induce a hierarchy in complexity over synthetic generations from the same document. 

\section{Data efficiency measurement}\label{app:data-eff-measurement}

We contextualize our loss improvements by estimating the data efficiency improvement or how much additional data our baseline would need to achieve the same loss improvement.

We estimate the data efficiency using the same methodology and scaling law as \cite{kim2025pre}, more specifically the reference-based strategy used in \cite{kotha2026replayingpretrainingdataimproves}. We first take the data scaling law for the standard recipe where the loss of the standard recipe under infinite compute is predicted to be $\frac{1.30}{D^{0.23}} + 1.89$. To compare two recipes, we measure the ``effective data'' of each recipe by inverting the scaling law (i.e. finding the value of $D$ that matches the loss). We then compute the data efficiency improvement as the ratio of the two effective data amounts. Since all three data scaling laws in the previous paper shared similar exponents and asymptotes and relative efficiency is determined by these two values, our results remain similar independent of the reference law. 

We highlight the two main differences between the data setting of this work and prior work.
\begin{enumerate}
    \item We are using a different subset of 200M DCLM tokens from the original paper. Specifically, instead of performing concat-and-chunk on the whole corpus and then selecting 200M tokens, this paper selects 164,000 DCLM documents. This slightly changes the sampling distribution since we are subsampling by documents and not tokens.
    \item All of our main body experiments in this paper use full cross-document attention while the previous paper masked cross-document attention. 
\end{enumerate}
We are not concerned about these changes since data efficiency is fully determined by the exponent and asymptote which should remain relatively stable.

\section{Additional evaluations}\label{app:additional-evals}

\subsection{Long-context evaluations}\label{app:additional-lc}

\begin{figure}
    \centering
    \includegraphics[width=0.45\textwidth]{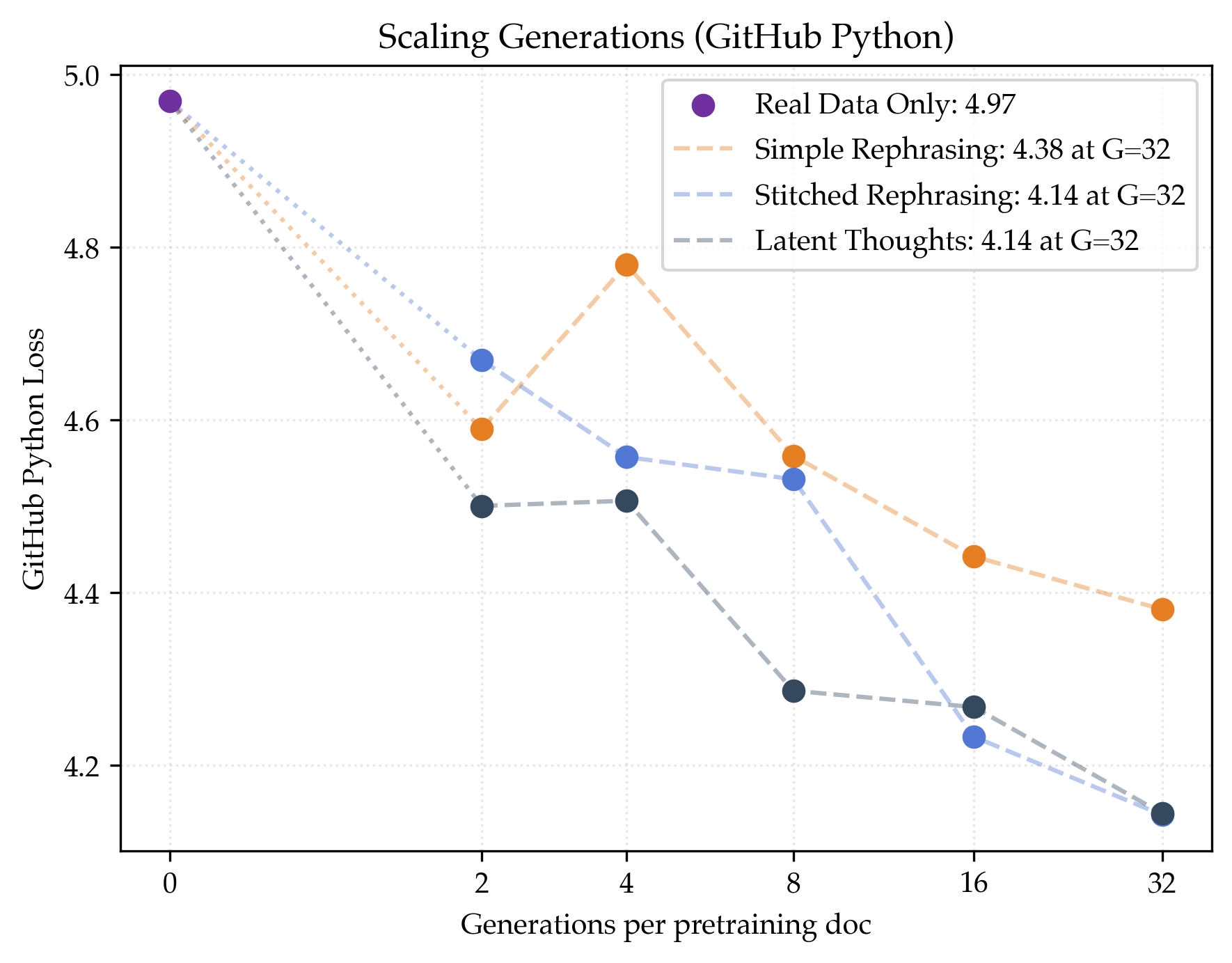}
    \includegraphics[width=0.45\textwidth]{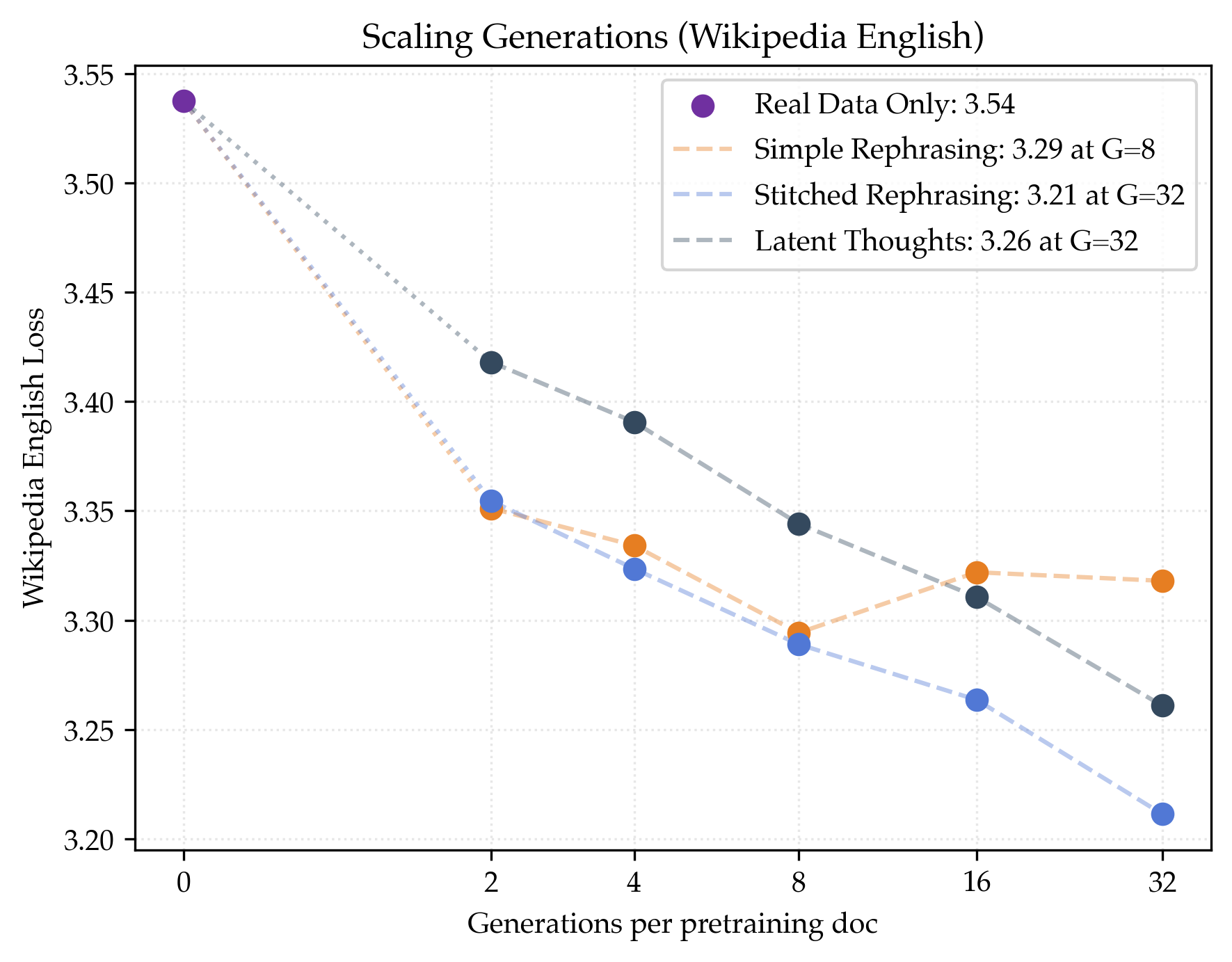}
    \caption{\textbf{Additional long-context evaluations.} Megadocs achieve lower loss and show less signs of plateauing than simple rephrasing on additional long-context evaluations on Github Python code and Wikipedia articles.}
    \label{fig:add_lc_evals}
\end{figure}

 Beyond arXiv papers, we measure loss on other natural long-context domains including English Wikipedia articles and Python code from Github \citep{uncheatable_eval}. All hyperparameters were selected via locally optimal i.i.d.~loss, making these evaluations truly held-out. In Figure \ref{fig:add_lc_evals}, we find that megadocs have better scaling trends than simple rephrasing. At the largest synthetic data scale of 32 generations per pre-training doc, our best megadoc method achieves a 0.24 loss improvement on Github Python and a 0.08 improvement on Wikipedia English over simple rephrasing.  

\subsection{Short vs. long-context evaluation}\label{app:short-vs-long}

We further investigate which data distributions megadocs provide an improvement on. We take the validation set (1000 documents) we use in our experiments and split it in half based on the tokenized document lengths. The ``short'' half contains 500 documents with a median length of $304$ tokens and a max length of $621$ tokens, while the ``long'' half contains 500 documents with a median length of $1230$ tokens and a max length of $36319$ tokens. 

\begin{figure}
    \centering
    \includegraphics[width=0.45\textwidth]{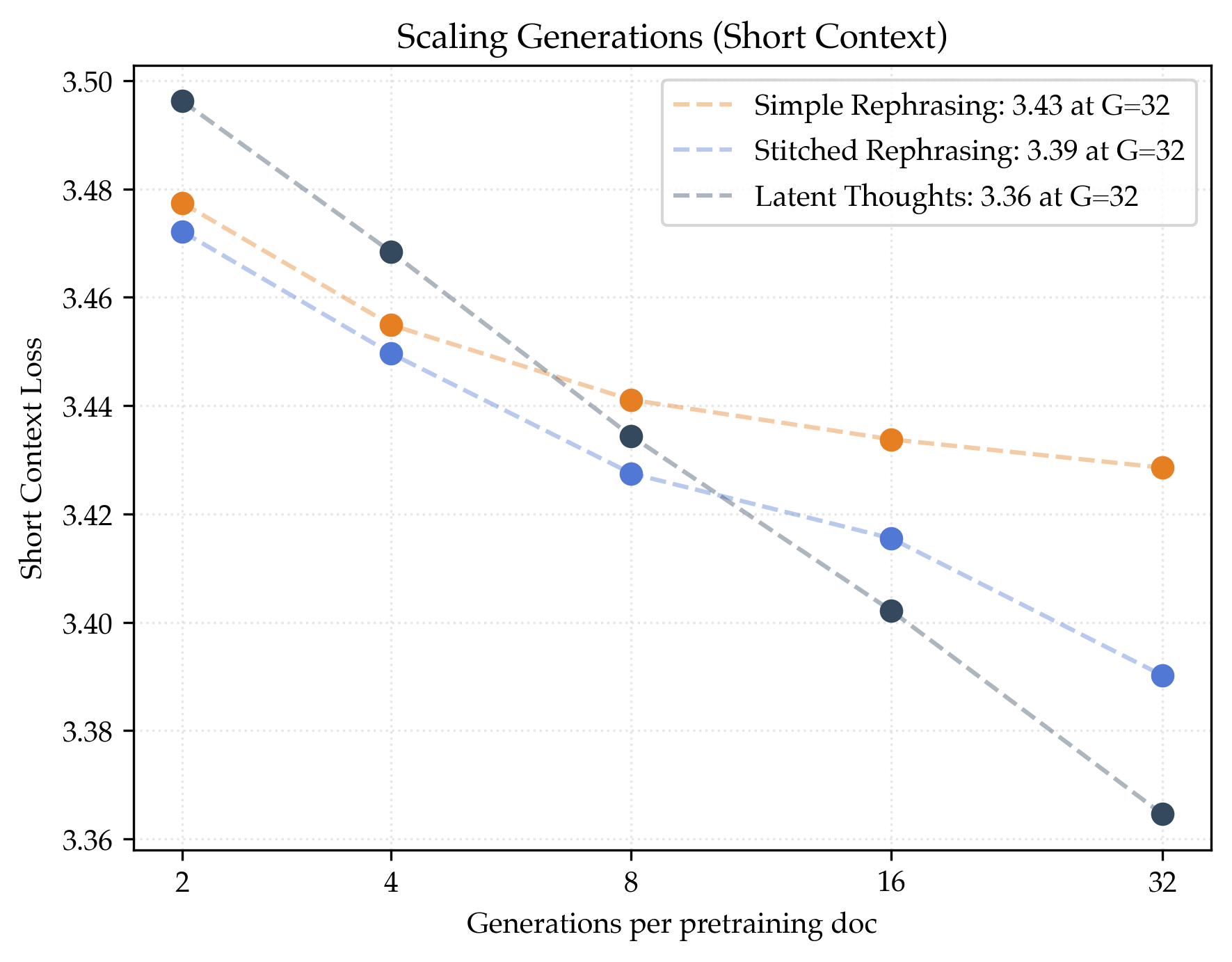}
    \includegraphics[width=0.45\textwidth]{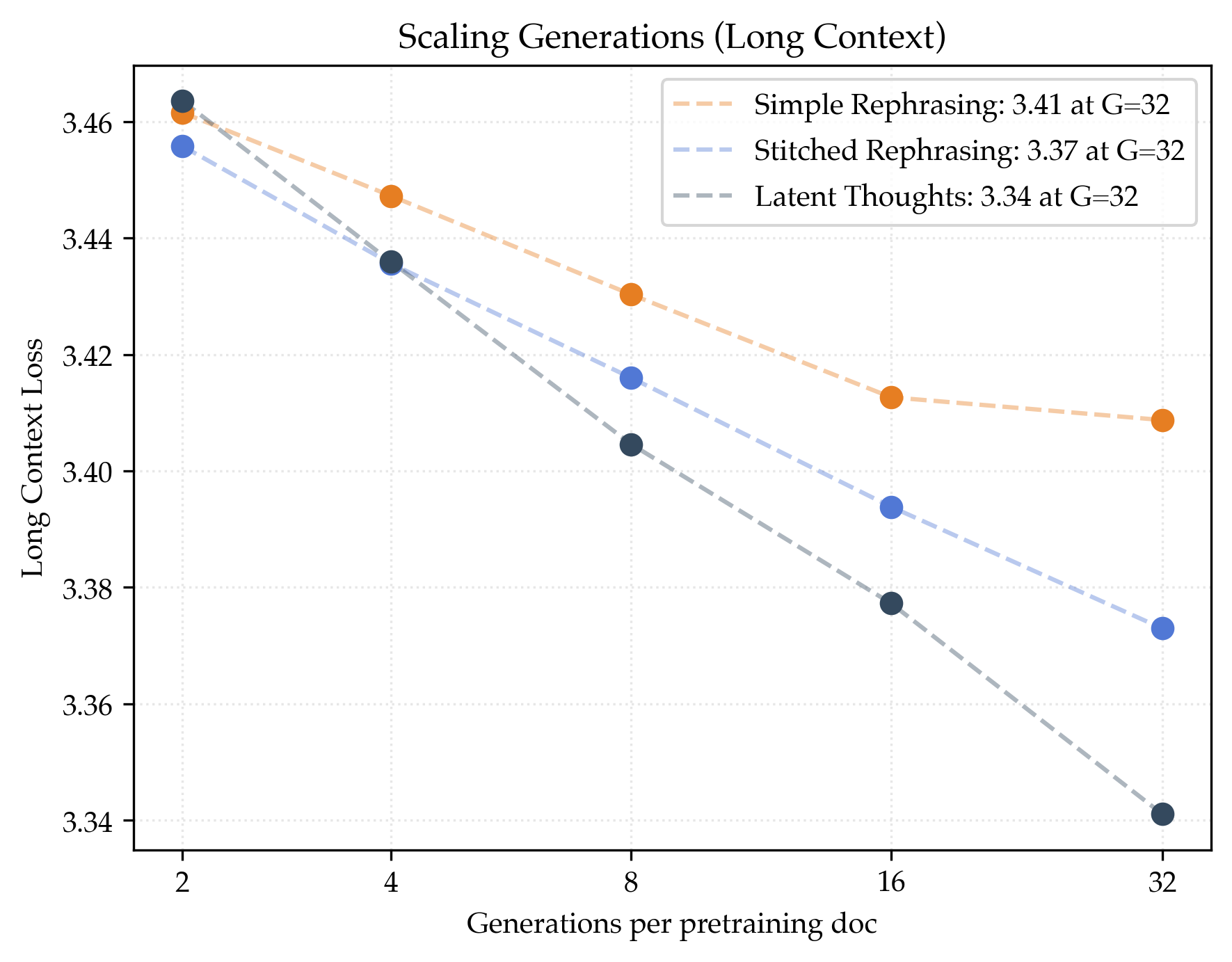}
    \caption{\textbf{Scaling trends on shorter vs longer documents.} We split the validation set in half (based on document length) and evaluate the scaling of our methods on both splits. The short and long splits have max document lengths of 621 and 36319 tokens respectively. The scaling of megadocs vs. simple rephrasing exhibits the same trend on the short and long splits, suggesting that megadocs help more generally with modeling many types of data.}
    \label{fig:short_long}
\end{figure}

We evaluate our models on both splits and surprisingly find that megadocs provide better scaling on both the short and long data splits (Figure \ref{fig:short_long}). This is additional evidence that megadocs don't just help for modeling long-context data but provide a better learning task with more general transfer.   

\section{Synthetic data generation details}\label{app:synth-gen-details}

\subsection{Generation procedure}
To create a synthetic data pool with $G$ generations per real document, we sample from Llama 3.1 8B Instruct $G$ times using the prompts in Appendix \ref{app:prompts} with temperature 1 using a high-throughput inference engine designed for batched
workloads \citep{juravsky2025tokasaurus}. For rephrasing and latent thoughts, we use max generation lengths of 1024 tokens and 512 tokens respectively. 

We adopt the exact same latent thoughts prompt as \cite{ruan2025reasoninglearnlatentthoughts}. When using the prompt, for each split point in a document, we use the entire prefix (before the split point) and suffix (after the split point) in the prompt. We note that this introduces more global information than the chunk-based splitting method in \cite{ruan2025reasoninglearnlatentthoughts}, but we don't ablate the effect of this choice.

\subsection{Synthetic data prompts}\label{app:prompts}

\textbf{System prompt:}
\begin{tcolorbox}
\small\ttfamily
Provide a direct response to the instructions without adding additional notes.
\end{tcolorbox}

\textbf{Rephrase prompt:}
\begin{tcolorbox}
\small\ttfamily
For the following document, regardless of its original content or formatting, write a full article of the same content in high quality English language as in texts on Wikipedia: \\ \\ 
Document:
\{doc\_text\}
\\ \\ 
Rephrased article:
\end{tcolorbox}

\textbf{Latent thoughts prompt:}
\begin{tcolorbox}
\small\ttfamily

You are provided with a pair of web document prefix and suffix. Your task is to insert latent thoughts
between them underlying the creation of the suffix conditioned on the prefix. The latent thoughts should
include any missing background knowledge and any reasoning traces underlying each claim (especially, step-by-step derivations or logical reasoning). \\ \\ 

Prefix: \{prefix\_text\} \\ \\ 

Suffix: \{suffix\_text\} \\ \\ 

Now provide the latent thoughts. Use concise, simple, and declarative language. Do not give any supporting
remarks or references to the terms 'prefix' and 'suffix', as this output will go directly into a computer program. Do not apply any markdown formatting or text embellishments. Optimize the content to ensure every word is informative, avoid vague language like 'xxx is essential'. Emphasize on the suffix without
repeating the content in the prefix. Focus on implicit reasoning and background knowledge that is not explicitly stated in the suffix, and use concrete logical reasoning or mathematical derivations when applicable.
\end{tcolorbox}

\section{Training details}\label{app:training-details}

\paragraph{Pre-training algorithm.} We instantiate the pre-training algorithm $\trainalg$ using a standard pre-training recipe developed through the Marin project (\url{https://marin.community}) by following the best practices shared publicly and found internally.

\begin{itemize}
    \item \textbf{Optimizer.} We train with AdamW, either with a default of $0.1$ or a tuned weight decay. We set other hyperparameters to standard defaults ($\beta_1 = 0.9, \beta_2 = 0.95, \epsilon = 10^{-8}$). We clip the norm of the gradient at $1$.
    \item \textbf{Learning rate.} We use a cosine learning rate schedule with a warmup for the first $1\%$ of training, decaying to 0 by the end of training. We always tune learning rate for all of our experiments. Every run has its own learning rate schedule and we never report the loss before the learning rate anneals to zero in the main body .
    \item \textbf{Architecture.} We train Llama-style auto-regressive language models. We specify the main architectural choices in Table~\ref{tab:model_configs}. When scaling models, we change the initialization scheme to have variance inversely proportional to the hidden dimension (this is known to outperform $\mu$P~\citep{yang2022tensorprogramsvtuning} within our framework: \url{https://github.com/marin-community/marin/issues/621}. For other architectural choices, we default to SiLU activations, untied word embeddings, and rotary position embeddings. 
    \item \textbf{Systems.} We train in mixed-precision with parameters in fp32 and compute + output in bf16. Most of our jobs were run on v4-64 TPUs, with bitwise-determinism for handling preemption and promoting reproducibility. 
    \item \textbf{Data.} We pre-train using DCLM  data~\citep{li2025datacomplmsearchgenerationtraining}. We keep a fixed validation set of 1000 sequences across all experiments for clear comparison. 
    \item \textbf{Data order.} We generate a random permutation of the windows after tokenization and use this same permutation across epochs. We note that loss might have been better if we used a unique permutation every epoch but keep this fixed across models which further reduces randomness of training.
\end{itemize}

\begin{table}[h]
\centering
\begin{tabular}{lrrrr}
\toprule
\textbf{Parameter} & \textbf{300M} & \textbf{1.5B} \\
\midrule
Context Length & 4096 & 4096  \\
Hidden Dimension & 512  & 1536 \\
Intermediate Dimension & 1792  & 5376 \\
Attention Heads & 8 & 16 \\
KV Heads & 8  & 8 \\
Layers & 6  & 36 \\
\bottomrule
\end{tabular}
\vspace{1em}
\caption{Model architecture configurations for different model sizes. We use the 1.5B model for student scaling experiments.}
\label{tab:model_configs}
\end{table}

\section{Relationship to ICPT and SBP}\label{app:sbp-icpt}

\subsection{Origin of megadocs}

We were initially interested in whether in-context pre-training (ICPT) \citep{shi2024incontextpretraininglanguagemodeling} was a data efficiency win, but we had trouble replicating the results of the paper at our small scale. We hypothesized that this was due to the ``density'' of the document embedding graph. The original paper operated at a scale of 300B pre-training tokens, where the number of related neighbors in the embedding graph was much higher than our setup of 200M tokens. 

We verified this hypothesis by doing ``density scaling'': we constructed pre-training datasets that had higher inter-document relationships by first sampling i.i.d. DCLM data at a higher token scale (e.g. 20B tokens) and then subsampling 200M tokens from the TSP ordering of this data. We were able to show that ICPT on denser data gives a large loss improvement on the corresponding validation set, and we found that the loss improvement scales with the density of the corpus. 

We then realized that synthetic rephrases were a ``free'' type of naturally dense data, which motivated the first version of stitching. We immediately observed large loss improvements on long-context loss, and we were able to distill these into i.i.d wins by mixing in the real data stream optimally. 

The success of stitching motivated the general concept of megadocs which led us to consider other forms of augmentation that stretch documents. This led to trying a simple version of Latent Thoughts \citep{ruan2025reasoninglearnlatentthoughts} which we found to be extremely effective.

\subsection{Stitched rephrasing as WRAP, ICPT, and SBP}

We speculate that stitched rephrasing gets the benefits of WRAP \citep{maini2024rephrasingwebrecipecompute}, ICPT \citep{shi2024incontextpretraininglanguagemodeling}, and SBP \citep{yang2025syntheticbootstrappedpretraining}.

We first note that SBP takes a document, retrieves its nearest neighbors in embedding space, and trains a generator conditioned on the original document. Note that if we dropped loss masking on the original document (which seems to have little effect in instruction following \citep{huertaenochian2024instructionfinetuningdoesprompt}), this is equivalent to 2 document ICPT. Therefore, we view ICPT as a multi-document generalization of SBP synthesizer training.

We then observe that SBP takes the generator and distills it into a model trained for i.i.d. loss. We hypothesize mixing the megadoc loss and real loss automatically serves the same purpose: the model learns features from the megadoc objective which transfer to the i.i.d. loss. Therefore, mixing stitched rephrasing automatically gets the benefits of SBP generalized to multiple documents as done in ICPT.

\end{document}